\documentclass{article}
\usepackage[utf8]{inputenc}

\usepackage{geometry}
\newgeometry{vmargin={1in}, hmargin={1.25in,1.25in}} 
\usepackage{microtype,soul}
\usepackage{graphicx}
\usepackage{subfigure}
\usepackage{booktabs} 
\usepackage{amsmath,amssymb,amsfonts,graphicx,nicefrac,mathtools,amsthm}
\usepackage{fancyhdr}
\usepackage{bbm}
\usepackage{tikz}
\usepackage{enumerate}
\usepackage{booktabs}
\usepackage{multirow}
\usepackage{xcolor}
\usepackage{enumitem}
\usepackage{makecell}

\usepackage{hyperref}

\def\pow{\mathcal{P}}
\def\R{\mathbb{R}}
\def\N{\mathbb{N}}

\def\C{\mathcal{C}^\omega}
\def\F{\mathcal{F}}

\def\Att{\text{A}}
\def\Int{\text{I}}

\def\I{\text{IG}}
\def\S{\Phi}
\def\Int{\text{I}}

\def\IG{\text{IG}}
\def\SP{\text{SP}}
\def\SP{\text{SP}}
\def\Shap{\text{Shap}}
\def\ST{\text{ST}}

\def\kth{k^{\text{th}}}
\def\pxi{\frac{\partial F}{\partial x_i}}

\def\pxii{\frac{\partial^2 F}{\partial x_i^2}}

\def\pxij{\frac{\partial^2 F}{\partial x_i \partial x_j}}

\def\IH{\text{IH}}

\newcommand{\specialcell}[2][c]{%
  \begin{tabular}[#1]{@{}c@{}}#2\end{tabular}}

\theoremstyle{definition}

\newtheorem{theorem}{Theorem}

\newtheorem{corollary}{Corollary}
\newtheorem{proposition}{Proposition}

\newtheorem{definition}{Definition}

\title{Distributing Synergy Functions: Unifying Game-Theoretic Interaction Methods for Machine-Learning Explainability
}

\author{Daniel Lundstrom\thanks{lundstro@usc.edu, University of Southern California} 
\and Meisam Razaviyayn\thanks{razaviya@usc.edu, University of Southern California}}

\begin{document}
\maketitle

\begin{abstract}
Deep learning has revolutionized many areas of machine learning, from computer vision to natural language processing, but these high-performance models are generally ``black box." Explaining such models would improve transparency and trust in AI-powered decision making and is necessary for understanding other practical needs such as robustness and fairness. A popular means of enhancing model transparency is to quantify how individual input features contribute to model outputs (called attributions) and the magnitude of interactions between groups of inputs. A growing number of methods import concepts and results from game theory to produce attributions and interactions. This work presents a unifying framework for game-theory-inspired attribution and $\kth$-order interaction methods. We show that, given modest assumptions, a \underline{unique} full account of interactions between features, called synergies, is possible for continuous input features. We identify how various methods are characterized by their policy of distributing synergies. We also demonstrate that gradient-based methods are  characterized by their actions on monomials, a type of synergy function, and introduce unique gradient-based methods. We show that the combination of various criteria uniquely defines the attribution/interaction methods. Thus, the community needs to identify goals and contexts when developing and employing attribution and interaction methods.\footnote{This work was supported in part with funding from the USC-Meta Center for Research and Education in AI \& Learning (REAL@USC center).}
\end{abstract}

\section{Introduction}
Explainability has become an ever increasing topic of interest among the Machine Learning (ML) community. Various ML methods, including deep neural networks, have unprecedented accuracy and functionality, but their functions are generally considered ``black box" and unexplained. Without ``explaining" a model's workings, it can be difficult to troubleshoot issues, improve performance, guarantee accuracy, or ensure other performance criteria such as fairness.


 A variety of approaches have been employed to address the explainability issue of  neural networks. Taking the taxonomy of \cite{linardatos2020explainable}, some methods are universal in application (called model agnostic) \cite{ribeiro2016should}, while other are limited to specific types of models (model specific) \cite{binder2016layer}. Some model-specific methods are limited to a certain data type, such as image \cite{selvaraju2017grad} or tabular data \cite{ustun2016supersparse}. Some methods are global, i.e., they seek to explain a model's workings as a whole \cite{ibrahim2019global}, while others are local, explaining how a model works for a specific input \cite{zeiler2014visualizing}. Finally, some methods seek to make models that are intrinsically explainable \cite{letham2015interpretable}, while others, called post hoc, are designed to be applied to a black box model without explaining it \cite{springenberg2014striving}. These post hoc methods may seek to ensure fairness, test model sensitivity, or indicate which features are important to a model's prediction.

This paper focuses on the concept of attributions and interactions. \textit{Attributions} are local, post hoc explanibility methods that indicate which features of an input contributed to a model's output \cite{lundberg2017unified}, \cite{sundararajan2017axiomatic}, \cite{sundararajan2020many}, \cite{binder2016layer}, \cite{shrikumar2017learning}. \textit{Interactions}, on the other hand, are methods that indicate which groups of features may have interacted, producing effects beyond the sum of their parts \cite{masoomi2021explanations}, \cite{chen2022generalized}, \cite{sundararajan2020shapley}, \cite{janizek2021explaining}, \cite{tsai2022faith}, \cite{blucher2022preddiff}, \cite{zhang2021interpreting}, \cite{liu2020detecting}, \cite{tsang2020feature}, \cite{hamilton2021axiomatic}, \cite{tsang2020does}, \cite{hao2021self}, \cite{tsang2017detecting}, \cite{tsang2018neural}. 
A common and fruitful approach to attributions and interactions is to translate and apply results from game theoretic cost sharing \cite{shapley1971assignment}, \cite{aumann1974values}. This has the advantages of already having a well-developed theory and producing methods that uniquely satisfy identified desirable qualities.

This work utilizes a game theoretic viewpoint to analyze, unify, and extend existing attribution and interaction methods.
The \underline{contributions} of this paper are as follows:
\begin{itemize}
    \item This paper offers a method of analysis for attribution and $\kth$ order interaction methods of continuous-input models through the concept of synergy functions. We show that, given natural and modest assumptions, synergy functions give a unique accounting of all interactions between features. We also show any continuous input function has a unique synergy decomposition.
    \item We highlight how various (existing) methods are governed by rules of synergy distribution, and common axioms constrain the distribution of synergies. With this in mind, we highlight the particular strengths and weaknesses of established methods.
    \item We show that under natural continuity criteria, gradient-based attribution and interaction methods on analytic functions are \underline{uniquely characterized} by their actions on monomials. This collapses the question ``how should we define interactions on analytic functions" to the question ``how should we define interactions of a monomial?" We then give two methods that serve as potential answers to this question.
    \item We discuss the  goal-dependent nature of attribution and interaction methods. Based on this observation, we identify a method for producing new attributions and interactions.
\end{itemize}

\section{Background}

\subsection{Notation and Terminology}
Let $N = \{1,...,n\}$ denote the set of feature indices in a machine learning model (e.g. pixel indices in an image classification model). For $a$, $b\in\R^n$, let $[a,b] = \{x\in\R^n : a_i\leq x_i \leq b_i \text{ for all } i \in N\}$ denote the hyper-rectangle with opposite vertices $a$ and $b$. 
Let $F:[a,b] \mapsto \mathbb{R}$ denote a machine learning model taking an input data point $x\in[a,b]$ and outputting a real number. For example, $F(x)$ can be viewed as the output of a softmax layer (for a specific class) in a neural network classifier.
We denote the class of such functions by $\F(a,b)$, or $\F$ if $a$, $b$ may be inferred. 
Define a \textit{baseline attribution method} as:
\begin{definition}[Baseline Attribution Method]
A baseline attribution method is any function of the form $\Att(x,x',F):D\rightarrow\R^n$, where $D\subseteq [a,b]\times[a,b]\times \F$. \footnotemark
\end{definition}

\footnotetext{Some attribution and interaction methods also incorporate the internal structure of a model. We do not consider these here. In other words, the attribution methods we consider only depends on the function $F(\cdot)$ and does not depend on how this function is implemented in practice.}

Baseline attribution methods give the contribution of each feature in an input feature vector, denoted $x\in[a,b]$, to a function's output, $F(x)$, with respect to some baseline feature vector $x'\in[a,b]$.\footnotemark We denote a general baseline attribution by $\Att$, so that $\Att_i(x,x',F)$ is the attribution score of feature $x_i$ to $F(x)$, with respect to the baseline feature values $x'$. The definition allows for attributions with more restricted domains than $[a,b]\times[a,b]\times \F$ because baseline attributions may require conditions on $F$ or $x$ in order to be well defined. We will see a simple example of such conditions when we define Integrated Gradient method in section~\ref{subsec:AttributionAndIntMethods}.\footnotetext{As an example, the first proposed baseline for image inputs was a black image, which corresponds to the zero vector \cite{sundararajan2017axiomatic}. The question of an appropriate baseline generally depends on the data. See \cite{VisualizingBaselines} for a survey of baselines for image tasks.} 
For the purpose of this paper, all attribution methods are baseline attribution methods.

While attribution methods give a score to the contribution of each input feature, \textit{Interactions} give a score to a group of features based on the group's contribution to $F(x)$ beyond the contributions of each feature \cite{grabisch1999axiomatic}. For ease of reference, we may speak of a nonempty set $S\subseteq  N$ as being a group of features, by which we mean the group of features with indices  in $S$. Let $\pow_k = \{S\subseteq  N: |S| \leq k\}$ contain all subsets of $N$ of size $\leq k$. Then we can define a $\kth$-order baseline interaction method by:
\begin{definition}[$\kth$-Order Baseline Interaction Method]
A $\kth$-order baseline attribution method is any function of the form $\Int^k(x,x',F):D\rightarrow\R^{|\pow_k|}$, where $D\subseteq [a,b]\times[a,b]\times \F$. 
\end{definition}
$\kth$-order interaction methods are a sort of expansion of attributions, giving a contribution for each group of features in $\pow_k$. For some $S\in \pow_k$, the term $\Int^k_S(x,x',F)$ indicates the component of $\Int^k(x,x',F)$ that gives interactions among the group of features $S$. When speaking of interactions among a group of features, there are multiple possible meanings: marginal interactions between members of a group, total interactions among members of the group, and average interactions among members of the group. Loosely speaking, if we let $G_S$ be the interactions among the features of $S$ that are not accounted for by the interactions of sub-groups, then $G_S$ represents marginal interactions of features in $S$, $\sum_{T\subseteq S} G_T$ represents the total interactions of features in $S$, and $\sum_{T\subseteq S} \mu_T G_T$ represents average interactions of features in $S$, where $\mu_T$ is some weight function. This paper focuses on marginal interactions.

Using quadratic regression as an example, suppose $F(x_1,x_2,x_3) = 2x_1-3x_2 + x_1x_3 - 15$, $x = (1,1,1)$, $x' = (0,0,0)$. Then a $2^\text{nd}$-order baseline interaction method may report something like: $\Int_{\emptyset}(x,x',F) = -15$, $\Int_{\{1\}}(x,x',F) = 2$, $\Int_{\{2\}}(x,x',F) = -3$, and $\Int_{\{1,3\}}(x,x',F) = 1$, and the other interactions equal zero. 

It should be noted that $1^\text{st}$-order interactions with $\Int^1_\emptyset$ disregarded and baseline attributions have equivalent definitions. As with attributions, interactions may not be defined for all $(x,x',F)$. We denote the set of inputs where a given $\Int^k$ is defined by $D_{\Int^k}$, or $D_\Att$ with regard to attributions. As with attributions, all interactions are baseline $\kth$-order interactions for the purpose of this paper. We may drop $x'$ if the baseline is fixed, and also drop $x$, implying that some appropriate value is considered.

\subsection{Axioms}
The definitions provided in the previous subsection are extremely general and may lead to attribution functions that are not practical. To find practically-relevant attributions or interaction methods, the standard strategy is to identify certain axioms a method should satisfy. This guarantees a method has desirable properties and constrains the possible forms a method can take. In this subsection, we review the common axioms of attributions and interactions used in prior work \cite{grabisch1999axiomatic} \cite{sundararajan2020shapley}, \cite{sundararajan2020many}, \cite{tsai2022faith}, \cite{janizek2021explaining}, \cite{marichal1999chaining}, \cite{zhang2020game}. Axioms are only presented for interactions; they can be easily reformulated for attributions by setting $k=1$ and disregarding $\Int^1_\emptyset$, so that $\Int^1(x,x',F):D\rightarrow \R^n$.
%
%
%
\begin{enumerate}[start]
    \item \textbf{Completeness}: $\sum_{S\in  \pow_k, |S|>0} \Int^k_S(x,x',F) = F(x) - F(x')$ for all $(x,x',F) \in D_{\Int^k}$.
\end{enumerate}
\textit{Completeness} can be described as requiring the sum of all interactions to account for the change in function value between the input and the baseline. Completeness is sometimes called efficiency in the game-theoretic literature and derives from the concept of cost-sharing \cite{shapley1971assignment},\cite{sundararajan2017axiomatic}. In the cost-sharing problem, a cost function gives the cost of satisfying the demands of a set of agents \cite{shapley1971assignment}. A cost-share is a division of the cost among all agents, thus requiring the sum of all cost shares to equal the total cost. In attributions and interactions, requiring completeness grounds the meaning of the interaction values by requiring the method account for the total function value change $F(x)-F(x')$. Thus the value of $\Int^k_S(x,x',F)$ indicates that the marginal interaction between features in $S$ contributed to the function changing by $\Int^k_S(x,x',F)$.
\begin{enumerate}[resume*]
\item \textbf{Linearity}: If $(x,x',F)$, $(x,x',G) \in D_{\Int^k}$, $a,b\in \R$, then $(x,x',aF+bG)\in D_{\Int^k}$, and $\Int^k(x,x',aF+bG) = a\Int^k(x,x',F)+b\Int^k(x,x',G)$.
\end{enumerate}
%
\textit{Linearity} ensures that when a model is a linear combination of sub-models,  the interactions or attributions of the model is a weighted sum  of the interactions or attributions of the sub-models.

We say that a function $F\in\F$ does not vary in some feature $x_i$ if for any vector $x\in[a,b]$, $f(t) = F(x_1,..,x_{i-1},t,x_{i+1},...,x_n)$ is constant. This indicates that $F$ is not a function of $x_i$. On the contrary, if it is false to say that $F$ does not vary in $x_i$, then we say $F$ varies in $x_i$. If $F$ does not vary in $x_i$, we call $x_i$ a null feature of $F$.
\begin{enumerate}[resume*]
    \item \textbf{Null Feature}: If $(x,x',F)\in D_{\Int^k}$, $F$ does not vary in $x_i$, and $i\in S$, then $\Int^k_S(x,x',F) = 0$.
\end{enumerate}
\textit{Null Feature} asserts that there is no  marginal interaction among a group if one of the features has no effect. There may be interactions between subsets of~$S$ so long as they do not contain a null feature.\footnotemark

\footnotetext{Null feature is similar to dummy as stated in \cite{sundararajan2017axiomatic} and \cite{sundararajan2020shapley}.}

The three axioms above, completeness, linearity, and null features, are generally assumed in the literature on game-theoretic attributions and interactions. Besides these three, there are many other axioms (guiding principles) offered that generally serve one of two purposes: either they distinguish a method as unique, or they show that a method satisfies desirable qualities. Among them are symmetry \cite{sundararajan2020shapley}, symmetry-preservation \cite{sundararajan2017axiomatic}, \cite{janizek2021explaining}, \cite{sundararajan2020many}, interaction symmetry \cite{janizek2021explaining}, \cite{tsai2022faith}, interaction distribution \cite{sundararajan2020shapley}, binary dummy \cite{grabisch1999axiomatic},\cite{sundararajan2020shapley}, sensitivity (sometimes called sensitivity (a))\cite{sundararajan2017axiomatic}, \cite{sikdar2021integrated}, implementation invariance \cite{sundararajan2017axiomatic}, \cite{sundararajan2020shapley}, \cite{janizek2021explaining}, \cite{sikdar2021integrated}, set attribution \cite{tsang2020does}, non-decreasing positivity \cite{lundstrom2022rigorous}, recursive axioms \cite{grabisch1999axiomatic}, 2-Efficiency \cite{grabisch1999axiomatic}, \cite{tsai2022faith}, faithfulness\footnotemark \cite{tsai2022faith}, affine scale invaraince \cite{friedman2004paths}, \cite{sundararajan2020many}, \cite{xu2020attribution}, demand monotonicity \cite{sundararajan2020many}, proportionality \cite{sundararajan2020many}, and causality \cite{xu2020attribution}. Some of the above axioms, such as linearity or implementation invariance, are satisfied by many methods, but no one method satisfies all axioms. For example, Faith-Shap \cite{tsai2022faith} agrees with the Shapley-Taylor's \cite{sundararajan2020shapley} axioms up to a point, but while Shapley-Taylor posits interaction distribution to gain a unique method, Faith-Shap instead posits a formulation of faithfulness to gain a unique method.
\footnotetext{While not stated as an axiom,  ``faithfulness" was given as a desirable property and used to constrain the form of an interaction.}

There are natural limitation to this setup, as some attributions in the literature do not satisfy these definitions and axioms. For example, GradCAM \cite{selvaraju2017grad} does not use a baseline input, nor does Smoothgrad \cite{smilkov2017smoothgrad}. Many methods, such as Layer-Wise Relevance Propagation \cite{zeiler2014visualizing} or Deconvolutional networks \cite{springenberg2014striving}, do not attempt to satisfy completeness, so that the magnitude of the attributions is governed by some other principle. It may be possible for some methods to be adjusted to have a baseline by taking the difference in attributions between an input and a baseline, or to satisfy completeness by scaling all attributions by some proportion. While considering adjusted methods could conceivably lead to interesting results, the methods in question are not designed to fit into a game-theoretic context, and we omit analysis of methods that need adjustment to fit in the game-theoretic paradigm.

\subsection{Attribution and Interaction Methods}
\label{subsec:AttributionAndIntMethods}
There are several well-known attributions based on cost sharing. Before we introduce them, we first introduce a necessary notation. For given features $S\subseteq  N$ and assumed baseline $x'$, we define $x_S\in [a,b]$ by:
\begin{equation}
    (x_S)_i =
        \begin{cases}
            x_i & \text{if } i \in S\\
            x_i' & \text{if } i \notin S,
        \end{cases}
\end{equation}
where $x_i$ is the $i^\text{th}$ element of $x$ and $x_i'$ is the $i^\text{th}$ element of $x'$. One well known attribution method is the \textbf{Shapley Value} \cite{shapley1971assignment}, \cite{lundberg2017unified}, given as:
\begin{equation} \nonumber
    \Shap_i(x,F) = \frac{1}{n}\sum_{S\subseteq  N\setminus\{i\}} {n-1 \choose |S|}^{-1} (F(x_{S\cup \{i\}}) - F(x_S)),
\end{equation}
where ${n-1 \choose |S|} \triangleq \frac{(n-1)!}{(n-1-|S|)! (|S|)!}$ denotes the number of subsets of size $|S|$ of $n-1$ features. The Shapley value is a direct import of the famous Shapley value in cost sharing literature into the ML context; it is obtained by considering all possible ways in which a vector $x'$ can transition to $x$ by sequentially toggling each component from the baseline value $x_i'$ to $x_i$. Specifically, $\Shap_i(x,F)$ is the average function change of $F$ when $x_i'$ toggles to $x_i$, over all possible transition sequences. The Shapley value is an example of a \textit{binary features method}, meaning it only considers $F$ evaluated at the points $\{x_S : S\subseteq  N\}$; that is, points where each feature value is the input or baseline value. The Shapley value is well defined for all $(x,x',F)\in[a,b]\times[a,b]\times\F$ and thus there is no need for domain restriction.

Several $\kth$-order interactions that extend Shapley values have been proposed, all of which are binary feature methods \cite{grabisch1999axiomatic},\cite{tsai2022faith}. First, define $\delta_{S|T}F(x)= \displaystyle{\sum_{W\subseteq  S}}(-1)^{|S|-|W|}F(x_{W\cup T})$, which intuitively measures the marginal impact of including the features in $S$ when the features in $T$ are already present based on the inclusion-exclusion principle.
%
The \textbf{Shapley-Taylor Interaction Index} of order $k$ \cite{sundararajan2020shapley} is then given by:
\begin{equation}
    \ST_S^k(x,F) =
    \begin{cases}
        \frac{k}{n} \sum_{T\subseteq  N\setminus S} \frac{\delta_{S|T}F(x)}{{n-1 \choose |T|}} & \text{if } |S|=k\\
        \delta_{S|\emptyset}(F) & \text{if } |S|<k.
    \end{cases}
\end{equation}
Shapley-Taylor prioritizes interactions of order $k$ and its unique contribution is to satisfy the interaction distribution axioms, which is discussed in \ref{section:synergies and axioms}.

Another well known attribution is the \textbf{Integrated Gradients} (IG) \cite{sundararajan2017axiomatic}:
\begin{equation}
    \IG_i(x,F) = (x_i-x_i') \int_0^1 \pxi(x' + t(x-x')) dt.
\end{equation}
The IG is a direct translation of the well known cost-sharing method of Aumann-Shapley~\cite{aumann1974values} to ML attributions. The IG has been called the continuous version of the Shapley value, insofar as it 1) makes use of the gradient, unlike binary features methods, and 2) is restricted to inputs $(x,x',F)$  such that $F$ is integrable on the path $x'+t(x-x')$.\footnotemark 
\, For the theoretical foundations of IG, see \cite{sundararajan2017axiomatic}, \cite{aumann1974values}, \cite{lundstrom2022rigorous}.
\footnotetext{For example, IG is not defined for $F(x_1,x_2) = \max(x_1,x_2)$ with $x'=(0,0)$, $x = (1,1)$}
Currently, no $\kth$-order interactions extension of the Integrated Gradient has been proposed. However, a 2-order interaction, \textbf{Integrated Hessian} (IH), has been proposed in \cite{janizek2021explaining}. This interaction method computes the pairwise interaction between $x_i$ and $x_j$ as:
\begin{equation} \nonumber
    \IH_{\{i,j\}}(x,F) =2(x_i-x_i')(x_j-x_j') \times \int_0^1\int_0^1 st\pxij(x'+st(x-x'))dsdt.
\end{equation}

The ``main effect" of $x_i$, or lone interaction (a misnomer), is defined as:
\begin{equation} \nonumber
    \begin{split}
        \IH_{\{i\}}(x,F) = &(x_i-x_i')\times \int_0^1\int_0^1\pxi(x'+st(x-x'))dsdt \\
        &+ (x_i-x_i')^2 \times \int_0^1\int_0^1st\pxii(x'+st(x-x'))dsdt.
    \end{split}
\end{equation}
IH is what we label a $\textit{recursive}$ $\textit{method}$ since  it is defined by using an attribution method recursively. Specifically, $\IH_{\{i,j\}}(x,F) = \IG_i(x,\IG_j(\cdot,F)) + \IG_j(x,\IG_i(\cdot,F))$. Similarly, $\IH_{\{i\}}(x,F) = \IG_i(x,\IG_i(\cdot,F))$ \cite{janizek2021explaining}. Note that the domain where IH is well defined is restricted to functions where components of the Hessian, $\text{H}_f = \nabla^2 F(x)$, can be integrated along the path $x' + t(x-x')$. We discuss the expansion of IH to a $\kth$-order interaction and its properties in section \ref{section:IH} and appendix \ref{IH_appendix}.

\subsection{The M\"obius Transform}\label{section:The Mobius Transform}
Lastly, we review the M\"obius transform, which will be useful for our definition of the notion of ``pure interactions" in section \ref{section:mobius transform as complete account of interactions}. Let $v$ be a real-valued function on $|N|$ binary variables, so that $v:\{0,1\}^N\rightarrow \R$. For $S\subseteq  N$, we write $v(S)$ to denote $v((\mathbbm{1}_{1\in S},...,\mathbbm{1}_{n\in S}))$, where $\mathbbm{1}$ is the indicator function. Recall that the M\"obius transform  of $v$ is a function $a(v):\{0,1\}^N\rightarrow \R$ given by \cite{rota1964foundations}:

\begin{equation}
    a(v)(S) = \sum_{T\subseteq  S} (-1)^{|S|-|T|} \, v(T).
\end{equation}
The M\"obius transform satisfies the following relation to $v$:

\begin{equation}\label{Mobius Identity}
        v(S) = \sum_{T\subseteq  N} a(v)(T)\mathbbm{1}_{T\subseteq  S} = \sum_{T\subseteq S} a(v)(T).
\end{equation}

The M\"obius transform can be conceptualized as a decomposition of $v$ into the marginal effects on $v$ for each subset of $N$. Each subset of $S$ has its own marginal effect on the change in function value of $v$, so that $v(S)$ is a sum of the individual effects, represented by $a(v)(T)$ in Eq.~\eqref{Mobius Identity}. 
For example, if $N = \{1,2\}$, then for
\begin{equation} \nonumber
    v(S) = \begin{cases}
        \alpha & \text{if } S=\emptyset \\
        \beta & \text{if } S=\{1\} \\
        \gamma & \text{if } S=\{2\} \\
        \delta & \text{if } S=\{1,2\} 
    \end{cases}
 \hspace{1cm}\text{we have}\hspace{1cm}
    a(v)(S) =
    \begin{cases}
        \alpha & \text{if } S = \emptyset\\
        \beta - \alpha & \text{if } S = \{1\}\\
        \gamma -\alpha& \text{if } S = \{2\}\\
        \delta - \beta - \gamma + \alpha & \text{if } S = \{1,2\}
    \end{cases}
\end{equation}
%

%
%
\section{M\"obius Transforms as a Complete Account of Interactions}\label{section:mobius transform as complete account of interactions}

\subsection{Motivation: Pure Interactions}\label{section:motivate pure interactions}
In order to identify desirable qualities of an interaction method, it would be fruitful to answer the question: what sorts of function is a ``pure interaction" of features in $S$? Specifically, is $F(x_1,x_2,x_3)=x_1x_2$ a function of pure interaction between $x_1$ and $x_2$? This question is useful because if $F$ is a pure interaction of $x_1$ and $x_2$ (i.e. the only effects in $F$ is an interaction between $x_1$ and $x_2$), then naturally it ought to be that $\Int^2_S(x,F) =0$ for $S\neq\{1,2\}$. Indeed, to continue the example, suppose $F$ is a general function and we can decompose $F$ as follows:
\begin{equation} \nonumber
    F(x) = f_\emptyset + \sum_{1\leq i\leq 3} f_{\{i\}}(x_i) + \hspace{-2mm}\sum_{1\leq i<j\leq 3} \hspace{-2mm} f_{\{i,j\}}(x_i,x_j) + f_{\{1,2,3\}}(x),
\end{equation}
where $f_\emptyset$ is some constant, $f_{\{i\}}$ is pure main effect of $x_i$; $f_{\{i,j\}}$ gives pure pairwise interactions; and $f_{\{1,2,3\}}$ is pure interaction between $x_1$, $x_2$, and $x_3$. Assuming $\Int^2$ conforms to linearity, we would gain:
 \begin{equation} \nonumber
        \Int^2_S(x,F) = \sum_{|T|\leq 3} \Int^2_S(x,f_T) = \Int^2_S(x,f_S) + \Int^2_S(x,f_{\{1,2,3\}}),
 \end{equation}
by applying the above principle, namely $\Int^2_S(x,f_T) =0$ if $S\neq T$, $|T|\leq2$. That is, the $2^\text{nd}$-order interaction of $F$ for~$S$ would be a sum of $\Int^2_S$ acting on the pure interaction function for group $S$, written $f_S$, and $\Int^2_S$ acting on a pure interaction of size $3$. This would generalize to higher order interactions, so that:
 \begin{equation} \nonumber
        \Int^k_S(x,F) = \Int^k_S(x,f_S) + \sum_{T\subseteq N, |T|>k} \Int^k_S(x,f_T).
 \end{equation}
We would then have to determine what rules should govern $\Int^k_S(x,f_S)$, and $\Int^k_S(x,f_T)$, $|T|>k$.

\subsection{Unique Full-Order Interactions}
In the previous section we spoke intuitively regarding the notion of pure interaction; we now present a formal treatment. Let $\Int^n$ be a $n^\text{th}$-ordered interaction function, i.e., $\Int^n$ gives the interaction between all possible subsets of features. In addition to the axioms of completeness and null features above, we propose two modest axioms for such a function; first, we propose a milder form of linearity, which requires linearity only for functions that $\Int_S^n$ assign no interaction to. We weaken linearity in the interest of establishing the notion of pure interactions with minimal assumptions.
\begin{enumerate}[resume*]
    \item \textbf{Linearity of Zero-Valued Functions}: If $(x,x',G)$, $(x,x',F)\in {D_{\Int^n}}$, $S\subseteq N$ such that $\Int^n_S(x,x',G)=0$, then $\Int^n_S(x,x',F+G) = \Int^n_S(x,x',F)$.
\end{enumerate}
Before introducing the next axiom, we consider the meaning of the baseline, $x'$. In cost sharing, the baseline is the state where all agents make no demands \cite{shapley1971assignment}. If an agent makes no demands, there are no attributions, nor are there interactions with other players. Likewise, the original IG paper notes \cite{sundararajan2017axiomatic}:
\begingroup
\addtolength\leftmargini{-0.18in}
\begin{quote}
\it
``Let us briefly examine the need for the baseline in the definition of the attribution problem. A common way for humans to perform attribution relies on counterfactual intuition. When we assign blame to a certain cause we implicitly consider the absence of the cause as a baseline for comparing outcomes. In a deep network, we model the absence using a single baseline input."
\end{quote}
\endgroup

As with the cost sharing literature and \cite{sundararajan2017axiomatic}, we interpret the condition $x_i=x_i'$ to indicate that the feature $x_i$ is not present. With this observation, and recalling that $x_S$ denotes a vector where the components in $S$ are not fixed at the baseline values in $x'$, we present the next axiom:
\begin{enumerate}[resume*]
    \item \textbf{Baseline Test for Interactions ($k=n$)}: For baseline $x'$, if $F(x_S)$ is constant $\forall x$, then $\Int^n_S(x,x',F) = 0$.
\end{enumerate}
This axiom states that if every variable $\notin S$ is held at the baseline value, and the other variables $\in S$ are allowed to vary, but the function is a constant, then there is no interaction between the features of $S$. Why is this sensible? The critical observation is that a feature being at its baseline value indicates the feature is not present. If the features of $S$ have no effect when other features are absent, then the features of $F$ do not interact in and of themselves and their interaction measurement should be zero. 

It should also be clarified that our definition of interactions allows $F$ and $x'$ to be chosen separately. However, it is generally the case that a ML model will be trained on data which will inform the appropriate choice of baseline. It is possible that a model does not admit to a baseline representing the absence of features, in which case game-theoretic baseline attributions and interactions may be ill-suited as explanation tools. We proceed to discuss the case when $F$ has a baseline, and assume implicitly that $x'$ is chosen as the fitting baseline to $F$.
\begin{theorem}\label{unique n-interactions}
    There is a unique $n$-order interaction method with domain $[a,b]\times[a,b]\times\F$ that satisfies completeness, null feature, linearity of zero-valued functions, and baseline test for interactions ($n=k$).
\end{theorem}
Proof of Theorem~\ref{unique n-interactions} can be found in Appendix~\ref{appendix: proof of unique n-order interaction}. We turn to explicitly defining the unique interaction function satisfying the conditions in Theorem~\ref{unique n-interactions}. For a fixed $x$ and implicit $x'$, note that $F(x_S)$ is a function of $S$. This implies it can be formulated as a function of binary variables indicating whether each input component of $F$ takes value $x_i$ or $x_i'$. Thus we can take the M\"obius transform of $F(x_{(\cdot)})$, written as $a(F(x_{(\cdot)}))$. Now, if we evaluate the M\"obius transform of $F(x_{(\cdot)})$ for some $S$, given as $a(F(x_{(\cdot)}))(S)$, and allow $x$ to vary, then this is a function of $x$. Recall that $\pow_k = \{S\subset N : |S| \leq k\}$. Given a baseline $x'$, define the \textbf{synergy function} as follows:

\begin{definition}[Synergy Function]
For $F\in\F$, $S\in\pow_n$, and implicit baseline $x'\in [a,b]$, the synergy function $\phi:\pow_N\times\F\rightarrow \F$ is defined by the relation $\phi_S(F)(x) = a(F(x_{(\cdot)}))(S)$.
\end{definition}
We present the following example to help illustrate the synergy function: let $F(x_1,x_2) = a + bx_1^2 + c\sin{x_2} + dx_1x_2^2$, and suppose $x'=(0,0)$ are the baseline values for $x_1$ and $x_2$ that indicate the features are not present. The synergy for the empty set is the constant $F(x') = a$, indicating the baseline value of the function when no features are present. To obtain $\phi_{\{1\}}(F)$, we allow $x_1$ to vary but keep $x_2$ at the baseline, and subtract the value of $F(x')$. This gives us $\phi_{\{1\}}(F)(x) = a+bx_1^2 - a = bx_1^2$. If instead we allow only $x_2$ to vary, we get $\phi_{\{2\}}(F)(x)  = a+c\sin(x_2) - a = c\sin(x_2)$. Finally, if we allow both to vary and subtract of all the lower synergies, we get $\phi_{\{1,2\}}(F)(x) = dx_1x_2^2$.

With the definition of the synergy function, we turn to the following corollary:
\begin{corollary}\label{corollary: synergy function uniquely satisfies}
    The synergy function is the unique $n$-order interaction method that satisfies completeness, null feature, linearity of zero-valued functions, and baseline test for interactions ($n=k$). 
\end{corollary}
%
Proof of Corollary~\ref{corollary: synergy function uniquely satisfies} is relegated to Appendix~\ref{appendix: proof of synergy function uniqueness}. The properties of the synergy function stem from properties of the M\"obius transform. Specifically, because the synergy function is defined by the M\"obious Transform, it inherits many of its properties, including completeness, null feature, linearity of zero-valued functions, and baseline test for interactions ($n=k$). The primary precursor to the synergy function is the Harsanyi dividend \cite{harsanyi1963simplified}, which is like the M\"obius transform and is formulated for discrete-input settings. More recently, the Shapley-Taylor Interaction Index \cite{sundararajan2017axiomatic} takes the form of the M\"obius Transform when $k=n$, where Shapley-Taylor imposes symmetry and interaction distribution axioms. Likewise, Faith-Shap \cite{tsai2022faith} takes the form of the M\"obius Transform when $k=n$, where Faith-Shap primarily imposes a best-fit property dubbed faithfulness. The novelty of the synergy function is that, while previous works assumed $F$ to be a set function (as in section~\ref{section:The Mobius Transform}), the synergy function is a linear functional between continuous input functions. Consequently, Corollary~\ref{corollary: synergy function uniquely satisfies} is novel, not only because of the inclusion of baseline test for interactions $(k=n)$, but also because all axioms do not assume $F$ is a set function.

%
\subsection{Properties of the Synergy Function}
Given a function $F$, the synergy of a single feature $x_i$ is given by
\begin{equation} \nonumber
    \phi_{\{i\}}(F)(x) = F(x_{\{i\}}) - F(x'),
\end{equation}
and the pairwise synergy for features $x_i$ and $x_j$ is
\begin{equation}
    \begin{split}
        \phi_{\{i,j\}}(F)(x) &= F(x_{\{i,j\}}) - \phi_{\{i\}}(F)(x) - \phi_{\{j\}}(F)(x) - F(x')\\
         &= F(x_{\{i,j\}}) - F(x_{\{i\}})- F(x_{\{j\}}) + F(x').\\
    \end{split}
\end{equation}
In general, the synergy function for a group of features $S$ is
\begin{equation} \nonumber
    \begin{split}
        \phi_{S}(F)(x) &= F(x_S) - \hspace{-3mm}\sum_{T\subsetneq S, T\neq \emptyset} \hspace{-3mm} \phi_T(F)(x) -F(x')\\
         &= \sum_{T\subseteq S} (-1)^{|S|-|T|}\times F(x_T).\\
    \end{split}
\end{equation}

With this we can define the notion of a pure interaction. A \textit{pure interaction function} of the features $S$ is a function that 1) takes a value of 0 if any feature in $S$ takes its baseline value, and 2) varies and only varies in the features in $S$.\footnote{For the degenerate case where $S=\emptyset$, a pure interaction of the features of $S$ would be a constant function.} This is exactly what the synergy function accomplishes: either $\phi_S(F)(x)= 0$, or $\phi_S(F)(x)$ varies in exactly the features in $S$ and is 0 whenever $x_i = x_i'$ for any $i\in S$. More technically, define $C_S = \{F\in\F|F \text{ is a pure interaction function of } S\}$ to be the set of pure interactions of features $S$. Then we have the following corollary:
\begin{corollary}\label{corollary: decompose F into synergies}
Suppose an implicit baseline $x'\in[a,b]$ and let $F\in\F$, and $S$, $T\in \pow_n$. Then the following hold:
\begin{enumerate}
    \item Pure interaction sets are disjoint, meaning $C_S\cap C_T = \emptyset$ whenever $S\neq T$.
    \item $\phi_S$ projects $\F$ onto $C_S\cup\{0\}$. That is, $\phi_S(F)\in C_S\cup \{0\}$ and $\phi_S(\phi_S(F)) = \phi_S(F)$.
    \item For $\S_T\in C_T$, we have $\phi_S(\S_T) = 0$ whenever $S\neq T$.
    \item $\phi$ uniquely decomposes $F\in\F$ into a set of pure interaction functions on distinct groups of features. That is, there exists $\pow\subset\pow_n$ such that $F=\sum_{S\in \pow} \S_S$ where each $\S_S\in C_S$, only one such representation exists, and $\S_S = \phi_S(F)$ for each $S\in \pow$ while $\phi_S(F) = 0$ for each $S\in \pow_n\setminus \pow$.
\end{enumerate}
\end{corollary}
Proof of Corollary~\ref{corollary: decompose F into synergies} is relegated to Appendix~\ref{appendix: proof of synergy funciton properties}. For ease of notation, we will move forward assuming that if $x'$ is not stated, the implicit baseline value is $x'=0$ and is appropriate to $F$. We will also assume that the synergy functions $S$ is applied using the proper implicit baseline choice. Lastly, we denote $\S_S\in C_S$ to be a  pure interaction in $S$ as defined above, or what we may also call a ``synergy function" in $S$.

\subsection{Axioms and the Distribution of Synergies}\label{section:synergies and axioms}

Now that we have the notion of pure interactions by way of the synergy function, we comment on the interplay between axioms and synergy functions. First, we present a version of the baseline test for interactions which applies for $k\leq n$. The idea here is a generalization of the ($k=n$) case; that if $\Int^k$ is a $\kth$-order interaction and $\S_S$ is some pure interaction function with $|S|\leq k$, then $\Int^k(\S_S)$ should not report interactions for any set but $S$. We give this as an axiom:
\begin{enumerate}[resume*]
    \item \textbf{Baseline Test for Interactions ($k\leq n$)}: For baseline $x'$ and any synergy function $\Phi_S$ with $|S|\leq k$, if $T\subsetneq S$, then $\Int^k_T(\Phi_S) = 0$.
\end{enumerate}
This is a weaker version of the defining axiom of  Shapley-Taylor~\cite{sundararajan2020shapley}, which states:
\begin{enumerate}[resume*]
    \item \textbf{Interaction Distribution}: For baseline $x'$ and any synergy function $\Phi_S$, if $T\subsetneq S$ and $|T|<k$, then $\Int^k_T(\Phi_S) = 0$.
\end{enumerate}
The baseline test of interactions asserts that if a synergy function is for a group of at least size $k$, $\Int^k$ should not report interactions for any other group. The interaction distribution asserts the same, and adds the caveat that if the synergy function is for a group of size larger than $k$, it must be distributed only to groups of size $k$.

We now detail how some of these axioms can be formulated as constraints on the distribution of synergies.
\begin{enumerate}[leftmargin=*]
    \item Completeness: ensures that any method distributes a synergy among sets of inputs. Formally, for a synergy function $\S_S$, we may say that $\Int^k_T(x,\S_S) = w_T(x,\S_S)\times\S_S(x)$, where $w_T$ is some function satisfying $\sum_{T\subseteq \pow_k}w_T(x,\S_S) = 1$.
    \item Linearity: enforce that $\Int^k(F)$ is the sum of $\Int^k$  applied to the synergies of $F$. Formally, $\Int^k(F) = \sum_{T\subset \pow_k} \Int^k(\phi_T(F))$.
    \item Null Feature: enforces that $\Int^k$ only distributed $\S_S$ to groups $T\subseteq  S$.
    \item Baseline Test for Interaction($k\leq n$): enforces that $\S_S$ is not distributed to groups $T\subsetneq S$ when $|S|\leq k$.
    \item Interaction Distribution: enforces that $\S_S$ is not distributed to groups $T\subsetneq S$ when $|S|\leq k$, and is distributed only to groups of size $k$ when $|S|>k$.
    \item Symmetry\footnote{See appendix \ref{appendix:symmetry} for a statement of symmetry axiom.}: enforces that a synergy $\S_S$ be distributed equally among groups in the binary features case.
\end{enumerate}
\section{Binary Feature Methods and Synergies}\label{section:binary feature methods}
We now  discuss the role of the synergy function in axiomatic attributions/interactions.
\cite{harsanyi1963simplified}\footnotemark noticed that for a synergy function $\S_S$, the Shapley value is
\footnotetext{\cite{harsanyi1963simplified} observed Eq.~\eqref{Shapley Distribution of Synergies} and \eqref{Shapley Definition by Synergies} in the binary feature setting with M\"obius transforms. Here we state the continuous input form with synergy functions.}
\begin{equation}\label{Shapley Distribution of Synergies}
    \Shap_i(\S_S) =
        \begin{cases}
            \frac{\S_S(x)}{|S|} & \text{if } i \in S\\
            0 & \text{if } i \notin S.
        \end{cases}
\end{equation}
This means the Shapley value distributes the function gain from~$\S_S$ equally among all $i\in S$. Using the synergy representation of $F$ and  linearity of Shapley values, we get
\begin{equation}\label{Shapley Definition by Synergies}
    \Shap_i(F) = \sum_{S\subseteq  N \, s.t.\,  i\in S} \frac{\S_S(x)}{|S|}.
\end{equation}
Thus, the Shapley value can be conceptualized as distributing each synergy $\S_{\{i\}}$ to $x_i$ and distributing all higher synergies, $\S_S$ with $|S|\geq 2$, equally among all features in $S$, e.g., $\Shap(\S_{\{1,2,3\}}) = (\frac{\S_{\{1,2,3\}}}{3},\frac{\S_{\{1,2,3\}}}{3},\frac{\S_{\{1,2,3\}}}{3},0,...,0)$. Indeed the Shapley value is characterized by its rule of distributing the synergy function.

\begin{proposition}\textit{(\cite[Thm 1]{grabisch1997k})}
The Shapley value is the unique attribution that satisfies linearity and acts on synergy functions as in \eqref{Shapley Distribution of Synergies}.
\end{proposition}

For a synergy function $\S_S$, the Shapley-Taylor interaction index of order $k$ for a group of features $T\in\pow_k$ is given by:
\begin{equation}\label{eq:Shapley-taylor distribution}
    \ST^k_T(\S_S) = \begin{cases}
            \S_S(x) & \text{if } T=S\\
            \frac{\S_S(x)}{{|S| \choose k}} & \text{if } T\subsetneq S, |T|=k\\
            0 & \text{else}
        \end{cases}.
\end{equation}
The Shapley-Taylor distributes each synergy function of $S$ to its group, unless is too large ($|S|>k$), in which case it distributes the synergy equally among all subsets of $S$ of size $k$. We denote this type of $\kth$-order interaction \textit{top-distributing}, as it projects all synergies larger than the largest available size, $k$, to the largest groups available. This results in Shapley-Taylor emphasising interactions between features of size $k$, which may be an advantage or disadvantage, depending on the goal of the interaction.

As with the Shapley value, the Shapley-Taylor is characterized by this action on synergy functions:

\begin{proposition} \textit{(\cite[Prop 4]{sundararajan2020shapley})}
The Shapley--Taylor Interatcion Index of order $k$ is the unique $\kth$-order interaction index that satisfies linearity and acts on synergy functions as in Eq.~\eqref{eq:Shapley-taylor distribution}.
\end{proposition}

There is another binary feature $\kth$-order interaction method similar to Shapley-Taylor, briefly motioned in \cite{sundararajan2020shapley}, with the distinction that it is not top-distributing. Here we detail and augment the method. Similarly to the Integrated Hessian, we may take the Shapley value recursively to gain pairwise interaction between $x_i$ and $x_j$, given by $\text{RS}_{\{i,j\}}(x,F) = \Shap_i(x,\Shap_j(\cdot,F))+\Shap_j(x,\Shap_i(\cdot,F)) = 2\Shap_i(x,\Shap_j(\cdot,F))$. Main effects for $x_i$ would be $\Shap_i(x,\Shap_i(\cdot,F))$.

More generally, consider expanding the expression $\|y\|_1^k$, and let $N^k_T$ denote the sum of coefficients associated exactly with the variables with indices in $T$. Then the {\textbf{Recursive Shapley}} of order $k$ distributes synergy functions as such:
\begin{equation}\label{eq: recursive shap distribution}
    \text{RS}^k_T(\S_S) = \begin{cases}
        \frac{N^k_T}{|S|^k}\S_S(x) & \text{if } T\subseteq S\\
        0 & \text{else}
    \end{cases}.
\end{equation}
where in the case $T=S=\emptyset$ we set $\frac{N_T^k}{|S|^k} := 1$. This formulation, however, has the disadvantage of distributing a portion of synergy functions for groups sized $\leq k$ to subgroups. For example, the recursively Shapley reports that a synergy function $\Phi_{\{1,2,3\}}(x)$ also has interactions for subgroup $\{1,2\}$. This violates the baseline test for interactions ($k\leq n$). We can modify the method to avoid this issue, causing Recursive Shapley to satisfy the baseline test for interactions ($k\leq n$) axiom. We explicitly detail the Recursive Shapley and modification in \ref{appendix:Recursive Shapley}. As with the Shapley method and Shapley-Taylor, we have the following Theorem:
\begin{theorem}\label{theorem: Recursive Shapley}
    The Recursive Shapley of order $k$ is the unique $\kth$-order interaction index that satisfies linearity and acts on synergy functions as in Eq.~\eqref{eq: recursive shap distribution}.
\end{theorem}
Proof of this Theorem can be found in Appendix~\ref{appendix: reursive shapley distribution}.
\section{Synergy Distribution in Gradient-Based Methods}
A critical aspect of the above binary feature methods is that they treat all features in a synergy function as equal contributors to the function output. For example, consider the synergy function of $S=\{1,2\}$ given by $F(x_1,x_2) = (x_1-x_1')^{100}(x_2-x_2')$. $F$ evaluated at $x=(x_1'+2,x_2'+2)$ yields $F(x) = 2^{100}2^1=2^{101}$. The Shapley method applied to $F$ treats both inputs as equal contributors, and would indicate that $x_1$ and $x_2$ each contributed $\frac{2^{101}}{2}$ to the function increase from the baseline. This assertion seems unsophisticated, not to mention intuitively incorrect, given we know the mechanism of the interaction function.

The IG exhibits the potential advantages of gradient-based attribution methods by providing a more sophisticated attribution. For $m\in \N^n$, define $(x-x')^m = (x_i-x_i')^{m_1}\cdot\cdot\cdot(x_n-x_n')^{m_n}$, taking the convention that if $m_i=0$ and $x_i=x_i'$, then $(x_i-x_i')^{m_i} = 1$. Define $m! = m_1!\cdot\cdot\cdot m_n!$, and define $D^m F = \frac{\partial^{\|m\|_1} F}{\partial x_1^{m_1}\cdot \cdot \cdot \partial x_n^{m_n}}$. We notate the non-constant features of $x^m$ by $S_m = \{i|m_i>0\}$.

We call a function of the form $F(y) = (y-x')^m$ a monomial centered at $x'$, and note that any monomial centered at an assumed baseline $x'$ is a synergy function of $S_m$. Assuming $m_i>0$ and taking $x'=0$, the IG attribution to $y^m$, a synergy function of $S_m$, is:

\begin{align}\label{IG_monomial_distribution}
    \IG_{\{i\}}(x,y^m) &= x_i\int_0^1 m_i(tx)^{(m_1,...,m_i-1,...,m_n)} dt \nonumber\\ 
    &= x_i\int_0^1 m_i t^{\sum m_i - 1} x^{(m_1,...,m_i-1,...,m_n)} dt\nonumber\\
    &= m_i x^m \frac{t^{\sum m_j}}{\sum m_j}\Big|_0^1\nonumber\\
    &= \frac{m_i}{\|m\|_1} x^m
\end{align}
This means that IG distributes the function change of $F(y) = y^m$ to $x_i$ in proportion to $m_i$. For example, the IG's attribution to our previous problem is $\IG((2,2),x_1^{100}x_2) = (\frac{100}{101}2^{101},\frac{1}{101}2^{101})$, a solution that seems much more equitable than the Shapley value. Thus the IG can distinguish between features based on the form of the synergy, unlike the Shapley value, which treats all features in a synergy functions as equal contributors.

\subsection{Continuity Condition}
We now move to more rigorously develop the connection between gradient-based methods and to monomials. To connect the action of attributions and interactions on monomials to broader functions, we now move towards defining the notion of an interaction being continuous in $F$. Let $\C$ denote the set of functions that are real-analytic on $[a,b]$. It is well known that any $F\in\C$ admits to a convergent multivariate Taylor Expansion centered at $x'$:

\begin{equation}
    F(x) = \sum_{m\in\N^n} \frac{D^mF(x')}{m!}(x-x')^m.
\end{equation}

Functions in $C^w$ have continuous derivatives of all orders, and those derivatives are bounded in $[a,b]$. Thus, $C^\omega$ it is a well-behaved class that gradient-based interactions ought to be able to assess.

Recall that the Taylor approximation of order $l$ centered at $x'$, denoted $F_l$, is given by:

\begin{equation}
    T_l(x) = \sum_{m\in\N^n, \|m\|_1\leq l} \frac{D^m(F)(x')}{m!}(x-x')^m.
\end{equation}

The Taylor approximation for analytic functions has the property that $D^mT_l$ uniformly converges to $D^mF$ for any $m\in\N^n$ and $x\in[a,b]$. Given this fact, it would be natural to require that for a given $\kth$-ordered interaction $\Int^k$ defined for $C^w$ functions, $\lim_{l \rightarrow \infty} \Int^k(T_l) = \Int^k(F)$.

This notion is further justified by the fact that many ML models are analytic. Particularly, NNs composed of fully connected and residual layers, analytic activation functions such as sigmoid, mish,  swish, as well as softmax layers are real-analytic. While models using max or ReLU functions are not analytic, they can be approximated to arbitrary precision by analytic functions simply by replacing ReLU and max with the parameterized softplus and smoothmax functions, respectively.

With this, we propose a continuity axiom requiring interactions for a sequence of Taylor approximations of $F$ to converge to the interactions at $F$.

\begin{enumerate}[resume*]
    \item \textbf{Continuity of Taylor Approximation for Analytic Functions}: If $\Int^k$ is defined for all $(x,x',F)\in[a,b]\times[a,b]\times\C$, then for any $F\in\C$, $\lim_{l \rightarrow \infty}\Int^k(x,x',T_l) = \Int^k(x,x',F)$, where $T_l$ is the $l^\text{th}$ order Taylor approximation of $F$ centered at $x'$.
\end{enumerate}
%
From this we have the following result, who's proof can be found in Appendix~\ref{appendix: proof of continuity theorem for interactions}:
\begin{theorem}\label{theorem: continuity condition}
    Let $\Int^k$ be an interaction method defined on $[a,b]\times[a,b]\times\C$ which satisfies linearity and continuity of Taylor approximation for analytic functions. Then $\Int^k(x,x',F)$ is uniquely determined by the the values $\Int^k$ takes for the inputs in the set $\{(x,x',F): F(y) = (y-x')^m, m\in\N^n\}$.
\end{theorem}
%
In section~\ref{section:binary feature methods} we saw that binary feature methods distribute synergy functions according to a rule, and that rule characterized the method as a whole. gradient-based methods satisfying linearity and the continuity condition are characterized by their actions on specific sets of elementary synergy functions, monomials. Thus, given our the continuity condition and linearity, we have collapsed the question of continuous interactions to the question of interactions of monomials centered at $x'$. Specifically, if linearity and continuity are deemed desirable, and a means of distributing polynomials can be chosen, then the entire method is determined for analytic functions. This is illustrated by the following corollary, with accompanying proof located in Appendix~\ref{appendix: continuity condition theorem}:

\begin{corollary}\label{IG corollary}
    IG is the unique attribution method on analytic functions that satisfies linearity, the continuity condition, and acts on the inputs $(x,x',(y-x')^m)$ as in Eq.~\eqref{IG_monomial_distribution}.
\end{corollary}

\subsection{Integrated Hessians}\label{section:IH}
Next, we present two gradient-based interaction methods corresponding to Shapley-Taylor and Recursive Shapley.

For $m\in\N^n$, the Integrated Hessian of $F(y) = y^m$ at $x$ is:

\begin{equation} \nonumber
    \begin{split}
        \IH_{\{i,j\}}(y^m) = \frac{2m_im_j}{\|m\|_1^2}x^m,\quad 
        \IH_{\{i\}}(y^m) = \frac{m_i^2}{\|m\|_1^2}x^m.
    \end{split}
\end{equation}

As in Recursive Shapley, IH distributes a portion of any pure interaction monomial to all nonempty subsets of features in $S_m$, breaking the baseline test for interactions($k\leq n$). For example, although $F(x_1,x_2,x_3) = x_1x_2$ is a synergy function of $S=\{1,2\}$, IH distributes some of $F$ to main effects. This can be remedied by directly distributing single and pairwise synergies, then using IH to distribute monomials involving 3 or more variables. This augmented IH is given below:

\begin{equation} \nonumber
    \begin{split}
        \IH^*_T(x,F) &= \phi_T(F)(x) + \IH_T(x,F - \hspace{-3mm}\sum_{|S|\leq 2}\hspace{-2mm}\phi_S(F)).
    \end{split}
\end{equation}

Both IH and augmented IH can be extended to $\kth$-ordered interactions to produce a monomial distribution scheme. Consider the expansion of $\|m\|_1^k$, and let $M^k_T(m)$ denote the sum of the terms of the expansion involving exactly the $m_i$ where $i\in T$. Explicitly,

\begin{equation}
    M^k_T(m) =\sum_{l\in\N^n,|l| = k, S_l = T} {k \choose l}m^l.
\end{equation}

The augmented IH of order $k$ acts on monomial functions as follows:

\begin{equation}\label{IHk* Distribution Rule}
    \IH^{k*}_T(y^m) =
        \begin{cases}
            x^m & \text{if } T=S_m\\
            \frac{M^k_T(m)}{\|m\|_1^k}x^m & \text{if } T\subsetneq  S_m, |S_m|>k\\
            0 & \text{else} 
        \end{cases}.
\end{equation}
To explain, $\IH^{k*}$ distributes all monomial synergies of size $\leq k$ to their groups, and distributes monomial synergies of size $>k$ to subgroups of $S_m$ in proportion to $M^k_T(m)$. IH and augmented IH also satisfy completeness and null feature. A full treatment of both is given in appendix \ref{IH_appendix}.
\begin{corollary}\label{IH* corollary}
    $\IH^{k*}$ is the unique attribution method on analytic functions that satisfies linearity, the continuity condition, and distributes monomials as in Eq.~\eqref{IHk* Distribution Rule}.
\end{corollary}
\subsection{Sum of Powers: A Top-Distributing Gradient-Based Method}

Previously we outlined a $\kth$-order interaction that was not top-distributing. Now we now present the distribution scheme for a gradient-based top-distributed $\kth$-order interaction we call $\textbf{Sum of Powers}$.\footnotemark \, We present only its action on monomials here, and detail the method in Appendix~\ref{appendix:sum_of_powers}. Sum of Powers distributes a monomial as such:
\begin{equation}\label{SP distribution scheme}
    \SP^k_T(y^m) =
    \begin{cases}
        x^m & \text{if } T=S_m\\
        \frac{1}{{|S_m|-1 \choose k-1}} \frac{\sum_{i\in T}m_i}{\|m\|_1}x^m & \text{if } T\subsetneq S_m, |T|=k\\
        0 & \text{else}
    \end{cases}.
\end{equation}
The highlight is that Sum of Powers satisfies completeness, null feature, linearity, continuity condition, baseline test for interactions, and is a top-distributing method. Particularly for $y^m$, $|S_m|>k$, Sum of Powers distributes $y^m$ only to top subgroups $T$, $T=k$, and in proportion to $\sum_{i\in T} m_i$. We present a corollary below; see Appendix~\ref{appendix:sum_of_powers} for full details.
\begin{corollary}\label{Sum Power corollary}
    Sum of Powers is the unique attribution method on analytic functions that satisfies linearity, the continuity condition, and distributes monomials as in Eq.~\eqref{SP distribution scheme}.
\end{corollary}
%
%
%
\section{Concluding Remarks}
The paradigm of synergy distribution is a useful concept for the analysis and development of attribution and interaction methods. First, it can point out weaknesses in existing methods such as the Integrated Hessian and indicate improvements, second, it can lead to new methods such as the Sum of Powers method, and last, it allows new characterization results based on synergy or monomial distribution. As seen in the comparison of Shapley Value vs Integrated Gradient, synergy distribution can play an important role implicitly even when not explicitly discussed in the literature. However, the application of this analysis tool does not settle the question, ``which method is best?" There exists conflicting groups of axioms and various combinations of them produce unique interactions. The choice of whether to use a top-distributing or recursively defined method, a binary features or gradient-based method, or some other method may vary with the goal. For example, top-distributed methods may be preferable when explicitly searching for strong interactions of size $k$, while an iterative approach may be preferable when seeking to emphasizing all interactions up to size $k$.

For problems with continuous inputs, gradient-based methods seem to offer a more sophisticated means of distributing
synergies, as they distinguish between features when they
distribute a synergy function. Here again, it is not clear if any
given method represents a clear ``winner" to distribute
monomials. We have presented one top-distributing and one recursive method, but it is unclear if these methods are best in class. For instance, perhaps a top-distributing method that distributes monomials by some softmax-weighted scheme is preferable to Sum of Powers. In order to find such methods, one may try to find a linear operator $L_S:\C\rightarrow\C$ where the continuity criteria apply and $L_S(y^m) = c_S(y,m)y^m$ for some desirable weighting function $c_S$.
Finding such linear operators could produce a variety of attribution and interaction methods. 

In the authors' opinion, the possibility of the existence of one ``best" method is improbable as various combinations of different axioms lead to the development of unique methods. Thus, choosing methods based on the context of the application seems a more logical approach. Indeed, the existence of unique methods with individual strengths is already studied in game-theoretic cost-sharing literature\footnotemark.
\footnotetext{See the Shaplye value vs Aumann-Shapley value vs serial cost for cost-sharing \cite{friedman1999three}, or the Shapley vs Banzhaf interaction indices \cite{grabisch1999axiomatic}.}

\section{Acknowledgements}
This work was supported in part by a gift from the USC-Meta Center for Research and Education in AI and Learning. The authors are grateful to Mukund Sundararajan for comments and discussions on developing an interaction extension to the Integrated Gradient method.

\clearpage

\clearpage 
\bibliography{ref_XAI}

\newcommand{\etalchar}[1]{$^{#1}$}
\begin{thebibliography}{HDWX21}

\bibitem[AS74]{aumann1974values}
Robert~J. Aumann and Lloyd~S. Shapley.
\newblock {\em Values of Non-Atomic Games}.
\newblock Princeton University Press, Princeton, NJ, 1974.

\bibitem[BML{\etalchar{+}}16]{binder2016layer}
Alexander Binder, Gr{\'e}goire Montavon, Sebastian Lapuschkin, Klaus-Robert
  M{\"u}ller, and Wojciech Samek.
\newblock Layer-wise relevance propagation for neural networks with local
  renormalization layers.
\newblock In {\em International Conference on Artificial Neural Networks},
  pages 63--71. Springer, 2016.

\bibitem[BVS22]{blucher2022preddiff}
Stefan Bl{\"u}cher, Johanna Vielhaben, and Nils Strodthoff.
\newblock Preddiff: Explanations and interactions from conditional
  expectations.
\newblock {\em Artificial Intelligence}, 312:103774, 2022.

\bibitem[CY22]{chen2022generalized}
Dangxing Chen and Weicheng Ye.
\newblock Generalized gloves of neural additive models: Pursuing transparent
  and accurate machine learning models in finance.
\newblock {\em arXiv preprint arXiv:2209.10082}, 2022.

\bibitem[FM99]{friedman1999three}
Eric Friedman and Herve Moulin.
\newblock Three methods to share joint costs or surplus.
\newblock {\em Journal of economic Theory}, 87(2):275--312, 1999.

\bibitem[Fri04]{friedman2004paths}
Eric~J Friedman.
\newblock Paths and consistency in additive cost sharing.
\newblock {\em International Journal of Game Theory}, 32(4):501--518, 2004.

\bibitem[GR99]{grabisch1999axiomatic}
Michel Grabisch and Marc Roubens.
\newblock An axiomatic approach to the concept of interaction among players in
  cooperative games.
\newblock {\em International Journal of game theory}, 28(4):547--565, 1999.

\bibitem[Gra97]{grabisch1997k}
Michel Grabisch.
\newblock K-order additive discrete fuzzy measures and their representation.
\newblock {\em Fuzzy sets and systems}, 92(2):167--189, 1997.

\bibitem[Har63]{harsanyi1963simplified}
John~C Harsanyi.
\newblock A simplified bargaining model for the n-person cooperative game.
\newblock {\em International Economic Review}, 4(2):194--220, 1963.

\bibitem[HDWX21]{hao2021self}
Yaru Hao, Li~Dong, Furu Wei, and Ke~Xu.
\newblock Self-attention attribution: Interpreting information interactions
  inside transformer.
\newblock In {\em Proceedings of the AAAI Conference on Artificial
  Intelligence}, volume~35, pages 12963--12971, 2021.

\bibitem[HLZ{\etalchar{+}}21]{hamilton2021axiomatic}
Mark Hamilton, Scott Lundberg, Lei Zhang, Stephanie Fu, and William~T Freeman.
\newblock Axiomatic explanations for visual search, retrieval, and similarity
  learning.
\newblock {\em arXiv preprint arXiv:2103.00370}, 2021.

\bibitem[ILMP19]{ibrahim2019global}
Mark Ibrahim, Melissa Louie, Ceena Modarres, and John Paisley.
\newblock Global explanations of neural networks: Mapping the landscape of
  predictions.
\newblock In {\em Proceedings of the 2019 AAAI/ACM Conference on AI, Ethics,
  and Society}, pages 279--287, 2019.

\bibitem[JSL21]{janizek2021explaining}
Joseph~D Janizek, Pascal Sturmfels, and Su-In Lee.
\newblock Explaining explanations: Axiomatic feature interactions for deep
  networks.
\newblock {\em J. Mach. Learn. Res.}, 22:104--1, 2021.

\bibitem[LHR22]{lundstrom2022rigorous}
Daniel~D Lundstrom, Tianjian Huang, and Meisam Razaviyayn.
\newblock A rigorous study of integrated gradients method and extensions to
  internal neuron attributions.
\newblock In {\em International Conference on Machine Learning}, pages
  14485--14508. PMLR, 2022.

\bibitem[LL17]{lundberg2017unified}
Scott~M Lundberg and Su-In Lee.
\newblock A unified approach to interpreting model predictions.
\newblock {\em Advances in neural information processing systems}, 30, 2017.

\bibitem[LPK20]{linardatos2020explainable}
Pantelis Linardatos, Vasilis Papastefanopoulos, and Sotiris Kotsiantis.
\newblock Explainable ai: A review of machine learning interpretability
  methods.
\newblock {\em Entropy}, 23(1):18, 2020.

\bibitem[LRMM15]{letham2015interpretable}
Benjamin Letham, Cynthia Rudin, Tyler~H. McCormick, and David Madigan.
\newblock {Interpretable classifiers using rules and Bayesian analysis:
  Building a better stroke prediction model}.
\newblock {\em The Annals of Applied Statistics}, 9(3):1350 -- 1371, 2015.

\bibitem[LSZ{\etalchar{+}}20]{liu2020detecting}
Zirui Liu, Qingquan Song, Kaixiong Zhou, Ting-Hsiang Wang, Ying Shan, and Xia
  Hu.
\newblock Detecting interactions from neural networks via topological analysis.
\newblock {\em Advances in Neural Information Processing Systems},
  33:6390--6401, 2020.

\bibitem[MHX{\etalchar{+}}21]{masoomi2021explanations}
Aria Masoomi, Davin Hill, Zhonghui Xu, Craig~P Hersh, Edwin~K Silverman,
  Peter~J Castaldi, Stratis Ioannidis, and Jennifer Dy.
\newblock Explanations of black-box models based on directional feature
  interactions.
\newblock In {\em International Conference on Learning Representations}, 2021.

\bibitem[MR99]{marichal1999chaining}
Jean-Luc Marichal and Marc Roubens.
\newblock The chaining interaction index among players in cooperative games.
\newblock In {\em Advances in Decision Analysis}, pages 69--85. Springer, 1999.

\bibitem[PS20]{VisualizingBaselines}
Su-In~Lee Pascal~Sturmfels, Scott~Lundberg.
\newblock Visualizing the impact of feature attribution baselines, 2020.

\bibitem[Rot64]{rota1964foundations}
Gian-Carlo Rota.
\newblock On the foundations of combinatorial theory i. theory of m{\"o}bius
  functions.
\newblock {\em Zeitschrift f{\"u}r Wahrscheinlichkeitstheorie und verwandte
  Gebiete}, 2(4):340--368, 1964.

\bibitem[RSG16]{ribeiro2016should}
Marco~Tulio Ribeiro, Sameer Singh, and Carlos Guestrin.
\newblock " why should i trust you?" explaining the predictions of any
  classifier.
\newblock In {\em Proceedings of the 22nd ACM SIGKDD international conference
  on knowledge discovery and data mining}, pages 1135--1144, 2016.

\bibitem[SBH21]{sikdar2021integrated}
Sandipan Sikdar, Parantapa Bhattacharya, and Kieran Heese.
\newblock Integrated directional gradients: Feature interaction attribution for
  neural nlp models.
\newblock In {\em Proceedings of the 59th Annual Meeting of the Association for
  Computational Linguistics and the 11th International Joint Conference on
  Natural Language Processing (Volume 1: Long Papers)}, pages 865--878, 2021.

\bibitem[SCD{\etalchar{+}}17]{selvaraju2017grad}
Ramprasaath~R Selvaraju, Michael Cogswell, Abhishek Das, Ramakrishna Vedantam,
  Devi Parikh, and Dhruv Batra.
\newblock Grad-cam: Visual explanations from deep networks via gradient-based
  localization.
\newblock In {\em Proceedings of the IEEE international conference on computer
  vision}, pages 618--626, 2017.

\bibitem[SDA20]{sundararajan2020shapley}
Mukund Sundararajan, Kedar Dhamdhere, and Ashish Agarwal.
\newblock The shapley taylor interaction index.
\newblock In {\em International conference on machine learning}, pages
  9259--9268. PMLR, 2020.

\bibitem[SDBR14]{springenberg2014striving}
Jost~Tobias Springenberg, Alexey Dosovitskiy, Thomas Brox, and Martin
  Riedmiller.
\newblock Striving for simplicity: The all convolutional net.
\newblock {\em arXiv preprint arXiv:1412.6806}, 2014.

\bibitem[SGK17]{shrikumar2017learning}
Avanti Shrikumar, Peyton Greenside, and Anshul Kundaje.
\newblock Learning important features through propagating activation
  differences.
\newblock In {\em International Conference on Machine Learning}, pages
  3145--3153. PMLR, 2017.

\bibitem[SN20]{sundararajan2020many}
Mukund Sundararajan and Amir Najmi.
\newblock The many shapley values for model explanation.
\newblock In {\em International conference on machine learning}, pages
  9269--9278. PMLR, 2020.

\bibitem[SS71]{shapley1971assignment}
Lloyd~S Shapley and Martin Shubik.
\newblock The assignment game i: The core.
\newblock {\em International Journal of game theory}, 1(1):111--130, 1971.

\bibitem[STK{\etalchar{+}}17]{smilkov2017smoothgrad}
Daniel Smilkov, Nikhil Thorat, Been Kim, Fernanda Vi{\'e}gas, and Martin
  Wattenberg.
\newblock Smoothgrad: removing noise by adding noise.
\newblock {\em arXiv preprint arXiv:1706.03825}, 2017.

\bibitem[STY17]{sundararajan2017axiomatic}
Mukund Sundararajan, Ankur Taly, and Qiqi Yan.
\newblock Axiomatic attribution for deep networks.
\newblock In {\em International Conference on Machine Learning}, pages
  3319--3328. PMLR, 2017.

\bibitem[TCL17]{tsang2017detecting}
Michael Tsang, Dehua Cheng, and Yan Liu.
\newblock Detecting statistical interactions from neural network weights.
\newblock {\em arXiv preprint arXiv:1705.04977}, 2017.

\bibitem[TCL{\etalchar{+}}20]{tsang2020feature}
Michael Tsang, Dehua Cheng, Hanpeng Liu, Xue Feng, Eric Zhou, and Yan Liu.
\newblock Feature interaction interpretability: A case for explaining
  ad-recommendation systems via neural interaction detection.
\newblock {\em arXiv preprint arXiv:2006.10966}, 2020.

\bibitem[TLP{\etalchar{+}}18]{tsang2018neural}
Michael Tsang, Hanpeng Liu, Sanjay Purushotham, Pavankumar Murali, and Yan Liu.
\newblock Neural interaction transparency (nit): Disentangling learned
  interactions for improved interpretability.
\newblock {\em Advances in Neural Information Processing Systems}, 31, 2018.

\bibitem[TRL20]{tsang2020does}
Michael Tsang, Sirisha Rambhatla, and Yan Liu.
\newblock How does this interaction affect me? interpretable attribution for
  feature interactions.
\newblock {\em Advances in neural information processing systems},
  33:6147--6159, 2020.

\bibitem[TYR22]{tsai2022faith}
Che-Ping Tsai, Chih-Kuan Yeh, and Pradeep Ravikumar.
\newblock Faith-shap: The faithful shapley interaction index.
\newblock {\em arXiv preprint arXiv:2203.00870}, 2022.

\bibitem[UR16]{ustun2016supersparse}
Berk Ustun and Cynthia Rudin.
\newblock Supersparse linear integer models for optimized medical scoring
  systems.
\newblock {\em Machine Learning}, 102:349--391, 2016.

\bibitem[XVS20]{xu2020attribution}
Shawn Xu, Subhashini Venugopalan, and Mukund Sundararajan.
\newblock Attribution in scale and space.
\newblock In {\em Proceedings of the IEEE/CVF Conference on Computer Vision and
  Pattern Recognition}, pages 9680--9689, 2020.

\bibitem[ZCCZ20]{zhang2020game}
Hao Zhang, Xu~Cheng, Yiting Chen, and Quanshi Zhang.
\newblock Game-theoretic interactions of different orders.
\newblock {\em arXiv preprint arXiv:2010.14978}, 2020.

\bibitem[ZF14]{zeiler2014visualizing}
Matthew~D Zeiler and Rob Fergus.
\newblock Visualizing and understanding convolutional networks.
\newblock In {\em European conference on computer vision}, pages 818--833.
  Springer, 2014.

\bibitem[ZXZ{\etalchar{+}}21]{zhang2021interpreting}
Hao Zhang, Yichen Xie, Longjie Zheng, Die Zhang, and Quanshi Zhang.
\newblock Interpreting multivariate shapley interactions in dnns.
\newblock In {\em Proceedings of the AAAI Conference on Artificial
  Intelligence}, volume~35, pages 10877--10886, 2021.

\end{thebibliography}
\bibliographystyle{alpha}

\clearpage

\appendix
\section*{Appendix}

\section{Table of Methods}
All listed methods satisfy completeness, linearity, null feature, and symmetry. All gradient-based methods satisfy the continuity condition. All interaction methods also satisfy baseline test for interactions ($k\leq n$) unless otherwise noted. We do not list interaction distribution, which is a combination of baseline test for interactions ($k\leq n$) and being top-distributing in the binary features scheme. 
\begin{center}
\begin{tabular}{ |c|c|c| } 
 \hline
 Name & Properties & Distribution Rule \\ 
 \hline\hline
 Synergy Function & \specialcell{unique $n^\text{th}$-order \\ interaction}  & 
 $\phi_T(\S_S) =\begin{cases}
     \Phi_S & \text{if } S=T\\
     0 & \text{if } S\neq T
 \end{cases}
 $
 \\
  \hline
 Shapley Value  & \specialcell{attribution method\\binary features} &
$ 
 \Shap_i(\S_S) =
 \begin{cases}
     \frac{\S_S}{|S|} & \text{if } i\in S\\
     0 & \text{if } i\notin S
 \end{cases}
 $
 \\
  \hline
  Integrated Gradient  & \specialcell{attribution method\\gradient-based} &
$ \IG_i(y^m) =
 \begin{cases}
     \frac{m_i}{\|m\|_1}x^m & \text{if } i\in S_m\\
     0 & \text{if } i\notin S_m
 \end{cases}$
 \\ 
   \hline
 Shapley-Taylor  & \specialcell{binary features \\ top-distributing} & 
 $\ST_T^k(\S_S) =\begin{cases}
     \S_S & \text{if } T=S\\
     \frac{\S_S}{{|S|\choose k}} & \text{if } T\subsetneq S, |T|=k\\
     0 & \text{else}
 \end{cases}$
 \\
\hline
 Sum of Powers  & \specialcell{gradient-based \\ top-distributing} & 
$\SP^k_T(y^m) =
    \begin{cases}
        x^m & \text{if } T = S_m\\
        \frac{1}{{|S_m|-1 \choose k-1}} \frac{\sum_{i\in T}m_i}{\|m\|_1}x^m & \text{if } T\subsetneq S_m, |T|=k\\
        0 & \text{else}
    \end{cases}$
 \\ 
 \hline
  Recursive Shapley & \specialcell{binary features\\iterative\\breaks baseline test} & 
  $\text{RS}^k_T(\S_S) = \begin{cases}
        \frac{N^k_T}{|S|^k}\S_S(x) & \text{if } T\subseteq S\\
        0 & \text{else}
    \end{cases}
    $
 \\
  \hline
  \specialcell{Augmented \\ Recursive Shapley} & \specialcell{binary features \\ iterative} & $\text{RS}^{k*}_T(\S_S) = \begin{cases}
            \S_S(x) & \text{if } T=S\\
            \frac{N^k_T}{|S|^k}\S_S(x) & \text{if } T\subsetneq S, |S|>k\\
            0 & \text{else}
        \end{cases}$
 \\ 
  \hline
  Integrated Hessian & \specialcell{gradient-based\\iterative\\breaks baseline test} & 
  $\IH^k_T(y^m) = \begin{cases}
            \frac{M^k_T(m)}{\|m\|_1^k} x^m & \text{if } T\subseteq  S_m\\
            0 & \text{else}\\
        \end{cases}$
 \\ 
  \hline
  \specialcell{Augmented \\ Integrated Hessian} & \specialcell{gradient-based \\ iterative} & $\IH^{k*}_T(y^m) =
        \begin{cases}
            x^m & \text{if } T=S_m\\
            \frac{M^k_T(m)}{\|m\|_1^k}x^m & \text{if } T\subsetneq  S_m, |S_m|>k\\
            0 & \text{else} 
        \end{cases}$
 \\ 
 \hline
\end{tabular}
\end{center}

\section{Statement of Symmetry Axiom}\label{appendix:symmetry}
Let $\pi$ be an ordering of the features in $N$. We loosely quote the definition of symmetry from \cite{sundararajan2020shapley}, altering the binary feature setting to a continuous feature setting:
\begingroup
\addtolength\leftmargini{-0.18in}
\begin{quote}
\it
\begin{enumerate}[resume*]
\item Symmetry Axiom: for all $F\in\F$, for all permutations $\pi$ on $N$:
\begin{equation}
    \Int^k_S(x,x',F) = \Int^k_{\pi S}(\pi x, \pi x', F\circ \pi^{-1}),
\end{equation}
where $\circ$ denotes function composition, $\pi S := \{\pi(i):i\in S\}$, and $(\pi x)_{\pi (i)} = x_{i}$.
\end{enumerate}
\end{quote}
\endgroup

This axioms implies that if we relabel the features, then interactions for the relabeled features will concur with interactions before relabeling. It requires that the domain, $[a,b]$, is closed under permutations of inputs, meaning it is of the form $[a_1,b_1]^n$.

\section{Proofs of Synergy Function Claims}
\subsection{Proof of Theorem~\ref{unique n-interactions}}\label{appendix: proof of unique n-order interaction}

\begin{proof}
Let $\Int$ be any $n$-ordered interaction that satisfies the given axioms, and let $x,x'\in[a,b]\times[a,b]$ be arbitrarily chosen. We assume that all interactions are taken with respect to input $x$ and baseline $x'$. For ease of notation, we define $F_S(x) = F(x_S)$ for $F\in\F(x,x')$.

For any nonempty $S \in \pow_n$, note that $\Int_S(F) = \Int_S(F-F_S+F_S)$. Note that $(F-F_S)(x_S)$ is constant. Thus, $\Int_S(F-F_S) = 0$ for any $S\in\pow_k$ by the baseline test for interaction. Thus, by linearity of zero-valued functions, we have established that $\Int_S(F) = \Int_S(F_S)$ for any $S\in\pow_k$.

We now proceed by strong induction:

$|S|=1$ case: Let $i\in N$ and choose $F\in\F$. Note that $F_{\{i\}}$ does not vary with any feature but $x_i$. This implies that for $S\neq \{i\}$, $\Int_{S}(F_{\{i\}}) = 0$ by null feature. By completeness, $\Int_{\{i\}}(F_{\{i\}}) = F_{\{i\}}(x)- F_{\{i\}}(x')$, and $\Int_{\{i\}}(F)$ is uniquely determined. Thus $\Int_S(F)$ is uniquely determined for $|S|=1$.

$|S|\leq k \Rightarrow |S| = k+1$ case: Suppose that for any $G\in\F[a,b]$ and any $S\subseteq \{1,...,n\}$ such that $|S|\leq k$, $\Int_S(G)$ is uniquely determined. Let $T\in\pow_n, |T| = k+1$, $F\in\F$. It has been established that $\Int_T(F) = \Int_T(F_T)$. Note that for all $S\subsetneq T$, we have $|S| \leq k$, so $\Int_S(F_T)$ is uniquely determined by the induction hypotheses. Since $F_T$ does not vary in each $x_i$ such that $i\notin T$, we have $\Int_S(F_T) = 0$ for $S\nsubseteq T$ by null feature. By completeness, $F_T(x)-F_T(x') = \sum_{S\subseteq \pow_k} \Int_S(F_T) = \sum_{S\subseteq T} \Int_S(F_T)$. Thus $\Int_T(F_T) =  F_T(x) - F_T(x') - \sum_{S\subsetneq T} \Int_S(F_T)$. Since $\Int_T(F) = \Int_T(F_T)$ equals the sum of uniquely determined terms, $\Int_T(F)$ is uniquely determined.
\end{proof}
\subsection{Proof of Corollary \ref{corollary: synergy function uniquely satisfies}}\label{appendix: proof of synergy function uniqueness}
We proceed to show the synergy function satisfies completeness, linearity, null feature, and baseline test for interactions ($k\leq n$).
\begin{proof}
\textbf{Completeness}: For any $v:\{0,1\}^n\rightarrow \R$, \cite[Appendix 7.1]{sundararajan2020shapley} shows that the M\"obius transform has the property that,

\begin{equation}
    v(T) = \sum_{S\subseteq  T} a(v)(S).
\end{equation}
Using this, observe,
\begin{equation}
    \begin{split}
        F(x') + \sum_{S\in\pow_n} \phi_S(F)(x) &= \sum_{S\subseteq  N} a(F(x_{(\cdot)}))(S)\\
        &= F(x_N)\\
        &= F(x),
    \end{split}
\end{equation}
which established completeness.
\newline
\newline
\textbf{Linearity of Zero-Valued Functions}: We simply establish $\phi$ is linear.

\begin{equation}
    \begin{split}
        \phi_S(cF+dG)(x) &= a(cF(x_{(\cdot)})+dG(x_{(\cdot)}))(S)\\
        &= \sum_{T\subseteq  S} (-1)^{|S|-|T|} \, \left[(cF(x_{(\cdot)})+dG(x_{(\cdot)}))(T)\right]\\
        &= c\sum_{T\subseteq  S} (-1)^{|S|-|T|} \, F(x_{(\cdot)})(T) + d\sum_{T\subseteq  S} (-1)^{|S|-|T|} \, G(x_{(\cdot)})(T)\\
        &= c\phi_S(F)(x) + d\phi_S(G)(x)
    \end{split}
\end{equation}
\newline
\newline
\textbf{Baseline Test for Interactions}: Suppose $F(x_S)$ is constant.

\begin{equation}
    \begin{split}
        \phi_S(F)(x) &= a(F(x_{(\cdot)}))(S)\\
        &= \sum_{T\subseteq  S} (-1)^{|S|-|T|} \, F(x_{T})\\
        &= \sum_{T\subseteq  S} (-1)^{|S|-|T|} \, F(x')\\
        &= F(x') \sum_{0\leq i \leq |S|} {|S| \choose i} (-1)^{|S|-i}\\
        &= 0
    \end{split}
\end{equation}
\newline
\newline
\textbf{Null Feature}: Suppose $F$ does not vary in some $x_i$ and $i\in S$. Then,

\begin{equation}\label{synergy Null feature proof}
    \begin{split}
        \phi_S(F)(x) &= a(F(x_{(\cdot)}))(S)\\
        &= \sum_{T\subseteq  S} (-1)^{|S|-|T|} \, F(x_{T})\\
        &= \sum_{T\subseteq  S, i\in T} (-1)^{|S|-|T|} \, F(x_{T}) + \sum_{T\subseteq  S, i \notin T} (-1)^{|S|-|T|} \, F(x_{T})\\
        &= \sum_{T\subseteq  S\setminus\{i\}} (-1)^{|S|-(|T|+1)} \, F(x_{T\cup\{i\}}) + \sum_{T\subseteq  S\setminus\{i\}} (-1)^{|S|-|T|} \, F(x_{T})\\
        &= -\sum_{T\subseteq  S\setminus\{i\}} (-1)^{|S|-|T|)} \, F(x_{T}) + \sum_{T\subseteq  S\setminus\{i\}} (-1)^{|S|-|T|} \, F(x_{T}) \\
        &=0
    \end{split}
\end{equation}
\end{proof}

\subsection{Proof of Corollary \ref{corollary: decompose F into synergies}}\label{appendix: proof of synergy funciton properties}
\begin{proof}We proceed in the order given in Corollary~\ref{corollary: decompose F into synergies}.
\newline
\newline
\textbf{1. Pure interaction sets are disjoint, meaning $C_S\cap C_T = \emptyset$ whenever $S\neq T$.}

Suppose $S$, $T\in\pow_n$ with $T\neq S$. We proceed by contradiction and suppose $F\in C_S\cup C_T$. WLOG $\exists i \in S\setminus T$, implying that $F$ varies in feature $i$ since $F$ is a synergy function of $S$, and $F$ does not vary in feature $i$, since $F$ is a synergy function of $T$. This is a contradiction. Thus $C_S\cap C_T= \emptyset$.
\newline
\newline
\textbf{2. $\phi_S$ projects $\F$ onto $C_S\cup\{0\}$. That is, $\phi_S(F)\in C_S\cup \{0\}$ and $\phi_S(\phi_S(F)) = \phi_S(F)$}

Let $F\in \F$. First, for the degenerate case, $\phi_\emptyset(F) = F(x')$, which is a constant function. For any constant $c$, $\phi_\emptyset(c) = c$, implying $\phi_\emptyset$ is a projection and surjective for the range $C_\emptyset\cup\{0\}$. Thus $\phi_\emptyset$ projects $\F$ onto $C_\emptyset \cup \{0\}$.

Now we will show that $\phi_S(F)$ either is a pure interaction of $S$ or is 0 in the non-degenerate case. Suppose $x_i=x_i'$ for some $i\in S$. Then,

\begin{equation} \nonumber
    \begin{split}
        \phi_S(F)(x) &= \sum_{T\subseteq  S} (-1)^{|S|-|T|} \, F(x_{T})\\
        &= \sum_{T\subseteq  S, i\in T} (-1)^{|S|-|T|} \, F(x_{T}) + \sum_{T\subseteq  S, i \notin T} (-1)^{|S|-|T|} \, F(x_{T})\\
        &= \sum_{T\subseteq  S\setminus\{i\}} (-1)^{|S|-(|T|+1)} \, F(x_{T\cup\{i\}}) + \sum_{T\subseteq  S\setminus\{i\}} (-1)^{|S|-|T|} \, F(x_{T})\\
        &= -\sum_{T\subseteq  S\setminus\{i\}} (-1)^{|S|-|T|)} \, F(x_{T}) + \sum_{T\subseteq  S\setminus\{i\}} (-1)^{|S|-|T|} \, F(x_{T}) \\
        &=0
    \end{split}
\end{equation}
Thus $\phi_S(F) = 0$ whenever $x_i=x_i'$ for some $i\in S$, and $\phi_S(F)$ satisfies condition 1 for being a pure interaction of $S$. 

Now, inspecting the definition, $\phi_S(F)(x) = \sum_{T\subseteq  S} (-1)^{|S|-|T|} \, F(x_{T})$, so $\phi_S(F)$ does not vary in $x_i$, $i\notin S$. Lastly, suppose that $F$ does not vary in some $x_i$, $i\in S$. Since $\phi$ satisfies null feature, $\phi_S(F) = 0$. So either $\phi_S(F)$ varies in all $x_i$ such that $i\in S$, or $\phi_S(F) = 0$. If the former, $\phi_S(F)$ satisfies condition 2 for being a pure interaction of $S$; if the latter, $\phi_S(F)=0$. Thus $\phi_S(F)=0$ or $\phi_S(F)$ is a pure interaction function of $S$, implying the range of $\phi_S$ is $C_S\cup\{0\}$.

Now let $\S_S\in C_S$. Note

\begin{align*}
    \phi_S(\S_S)(x) &= \sum_{T\subseteq  S} (-1)^{|S|-|T|} \, \S_S(x_{T})\\
    &= \sum_{T = S} (-1)^{|S|-|T|} \, \S_S(x_{T})\\
    &= \S_S(x_{S})\\
    &= \S_S(x)
\end{align*}
\newline
It is plain by the definition that $\phi_S(0) = 0$. Thus $\phi_S$ is surjective for the range $C_S\cup\{0\}$. Since the range of $\phi_S$ is $C_S\cup\{0\}$, $\phi$ maps elements of $C_S$ to themselves, and maps 0 to 0, so $\phi_S$ is a projection.
\newline
\newline
\textbf{3. For $\S_T\in C_T$, we have $\phi_S(\S_T) = 0$ whenever $S\neq T$.}

Let $\S_T\in C_T$ and $T\neq S$. If $\exists i \in S\setminus T$, then $\phi_S(\S_T)=0$ by null feature. Otherwise $S\subsetneq T$, and $\phi_S(\S_T) = 0$ be baseline test for interactions ($k=n$).
\newline
\newline
\textbf{4. $\phi$ uniquely decomposes $F\in\F$ into a set of pure interaction functions on distinct groups of features. That is, there exists $\pow\subset\pow_n$ such that $F=\sum_{S\in \pow} \S_S$, where each $\S_S\in C_S$. Further more, only one such representation exists, $\S_S = \phi_S(F)$ for each $S\in \pow$, and $\phi_S(F) = 0$ for each $S\in \pow_n\setminus \pow$.}

$F=\sum_{S\in\pow_n} \phi_S(F)$, and each $\phi_S(F)\in C_S\cup\{0\}$. Since $0+\phi_\emptyset(F) \in C_\emptyset$ and we may gather all the $\phi_S(F)$ terms that are zero into the $C_\emptyset$ term, we have shown a decomposition exists.

Let it be that $F(x) = \sum_{S\in\pow} \S_S(x)$ for some $\pow\in \pow_n$, where each $\S_S$ is an interaction function in $S$. By the results already established, we have for any $T\in \pow$

\begin{align*}
    \phi_S(F) &= \phi_S(\sum_{T\in\pow} \S_T)\\
    &= \sum_{T\in \pow} \phi_S(\S_T)\\
    &= \phi_S(\S_S)\\
    &= \S_S
\end{align*}

If $S\notin \pow$, then

\begin{align*}
    \phi_S(F) &= \phi_S(\sum_{T\in\pow} \S_T)\\
    &= \sum_{T\in \pow} \phi_S(\S_T)\\
    &= 0
\end{align*}

Now suppose that there are two decompositions, $\sum_{S\in \pow^1} \S^1_S = F = \sum_{S\in\pow^2}\S^2_S$. WLOG suppose $S\in\pow^1\setminus\pow^2$. Then $\phi_S(F) = 0$ since $S\notin \pow^2$ and $\phi_S(F) = \S^1_S$ since $S\in\pow^1$. Thus $\S_S^1 = 0$ and $S = \emptyset$. Thus $\pow^1 \triangle \pow^2$ equals either $\emptyset$ or $\{\emptyset\}$, and in the case that $\pow^1 \triangle \pow^2=\{\emptyset\}$ the extra term corresponding to $\emptyset$ in one of the sums is 0, and does not effect the decomposition. Now, if $\pow^1 \triangle \pow^2=\emptyset$, then for any $S\in \pow^1, \pow^2$, we have $\S_S^1 = \phi_S(F) = \S_S^2$. Thus, the decomposition is unique.
\end{proof}

\section{Interaction Methods}
Here we give an in depth treatment of the Recursive Shapley, Integrated Hessian, and Sum of Powers methods, as well as the augmentations to the recursive methods. We define the methods and show that each method is the unique method that satisfies linearity, their distribution policy, and in the case of gradient methods, the continuity condition. We also prove that each method satisfies desirable properties such as completeness, null feature, symmetry, and, if applicable, baseline test for interactions ($k\leq n$).

\subsection{Recursive Shapley and Augmented Recursive Shapley}\label{appendix:Recursive Shapley}
\subsubsection{Defining Recursive Shapley}
Here we detail the properties of Recursive Shapley and Augmented Recursive Shapley. Let $\sigma^k_T$ be the set of sequences of length $k$ such that the sequence is made of the elements of $T\neq \emptyset$ and each element appears at least once. For example, $\sigma_{\{1,2\}}^3 = \{(1,1,2),(1,2,1),(1,2,2),(2,1,1),(2,1,2),$ $(2,2,1)\}$. Calculating the size of $\sigma_T^k$, $|\sigma_T^k| = \sum_{|l|=k \text{ s.t. } S_l=T}{k\choose l}=N^k_T$. For a given sequence $s$, define $\IG_t(x,F)$ be a recursive implementation of the Shapley method according to the sequence $s$, i.e., $\Shap_{(1,2,3)}(x,F) = \Shap_3(x,\Shap_2(\cdot,\Shap_1(\cdot,F)))$. We can then define the $\kth$-order Recursive Shapley for $T\neq \emptyset$ as:

\begin{equation}
    \text{RS}^k_T(x,F) = \sum_{s\in\sigma^k_T}\Shap_s(x,F)
\end{equation}
and define $\text{RS}^k_\emptyset(x,x',F) := F(x')$.

We now move to inspect this equation and establish some properties. Eq.~\eqref{Shapley Distribution of Synergies} states that for a synergy function $\S_S$, $S\neq \emptyset$,
\begin{equation}
    \Shap_i(x,\S_S) =
        \begin{cases}
            \frac{\S_S(x)}{|S|} & \text{if } i \in S\\
            0 & \text{if } i \notin S
        \end{cases}
\end{equation}
Then for a given sequence $s\in\sigma^k_T$ and synergy function $\S_S$, if $T\subseteq  S$ then,

\begin{equation}
    \begin{split}
        \Shap_s(x,\S_S) &= \Shap_{s_k}(x,\Shap_{s_{k-1}}( ... \Shap_{s_1}(\cdot, \S_S)....)\\
        &= \Shap_{s_k}(x,\Shap_{s_{k-1}}(... \Shap_{s_2}(\cdot, \frac{\S_S}{|S|})....)\\
        &= \Shap_{s_k}(x,\Shap_{s_{k-1}}(... \Shap_{s_3}(\cdot, \frac{\S_S}{|S|^2})....)\\
        &= ...\\
        &= \Shap_{s_k}(x,\frac{\S_S}{|S|^{k-1}}))\\
        &= \frac{\S_S(x)}{|S|^k}\\
    \end{split}
\end{equation}
However, if $T\subsetneq S$ then there exists an element of $s$ that is not in $S$, and:

\begin{equation}
        \begin{split}
        \Shap_s(x,\S_S) &= 0,
    \end{split}
\end{equation}
due to some ${s_j}\notin S$ in the sequence.
\subsubsection{Recursive Shapley's Distribution Policy}\label{appendix: reursive shapley distribution}
Now, to show how Recursive Shapley \textul{distributes synergies}, apply the definition of recursive Shapely for $S\neq \emptyset$ to get:
\begin{equation}
    \begin{split}
        \text{RS}^k_T(x,\S_S) &= \sum_{s\in\sigma^k_T} \Shap_s(x,\S_S)\\
        &=\begin{cases}
            \sum_{s\in\sigma^k_T} \frac{\S_S(x)}{|S|^k} & \text{if } T\subseteq  S\\
            \sum_{s\in\sigma^k_T} 0 & \text{if } T\nsubseteq S\\
        \end{cases}\\
        &= \begin{cases}
            \frac{N^k_T}{|S|^k} \S_S(x) & \text{if } T\subseteq  S\\
            0 & \text{if } T\nsubseteq S\\
        \end{cases}
    \end{split}
\end{equation}
We also gain the above for $S=\emptyset$ by setting $\frac{N^k_T}{|S|^k}=1$ when $T=\emptyset$. This establishes the distribution scheme in Eq.~\eqref{eq: recursive shap distribution}.

Recursive Shapley is also \textul{linear} because it it the sum of function compositions of composition of linear functions. This establishes Theorem~\ref{theorem: Recursive Shapley}.

\subsubsection{Properties of Recursive Shapley}
To show Recursive Shapley satisfies \textul{completeness}, observe for $S\neq \emptyset$:

\begin{equation}
    \begin{split}
        \sum_{T\in\pow_k, |T|>0} \text{RS}_T^k(x,\S_S) &= \sum_{T\subseteq S} N^k_T \frac{\S_S(x)}{|S|^k} \\
        &= \frac{\S_S(x)}{|S|^k} \sum_{T\subseteq S} N^k_T\\
        &= \frac{\S_S(x)}{|S|^k} |S|^k\\
        &= \S_S(x)
    \end{split}
\end{equation}
The case when $S=\emptyset$ is easily verified by inspecting the synergy distribution policy of RS.

To show Recursive Shapley satisfies \textul{null feature}, suppose that $F$ does not vary in $x_i$. Then for any $S\in\pow_k$ such that $i\in S$, $\phi_S(F)=0$ since the synergy function is an interaction satisfying null feature. Then if $i\in T$,

\begin{equation}
    \begin{split}
        \text{RS}_T^k(x,F) &= \sum_{S\in\pow_k} \text{RS}_T^k(x,\phi_S(F))\\
         &= \sum_{S\in\pow_k \text{ s.t. } i\in S} \text{RS}_T^k(x,\phi_S(F)) + \sum_{S\in\pow_k \text{ s.t. } i \notin S} \text{RS}_T^k(x,\phi_S(F))\\
         &= \sum_{S\in\pow_k \text{ s.t. } i\in S} \text{RS}_T^k(x,0) + \sum_{S\in\pow_k \text{ s.t. } i \notin S} 0\\
         &= 0
    \end{split}
\end{equation}
Where the terms in the second sum are zero by Eq.~\eqref{eq: recursive shap distribution}.

To show Recursive Shapley satisfies \textul{symmetry}, let $\pi$ be a permutation on $N$. Note that for $\S_S\in C_S$, we have $\S_S\circ\pi^{-1}$ is a pure interaction function in $\pi S$ with baseline $\pi x'$. Then

\begin{align*}
    \text{RS}^k_{\pi T}(\pi x,\pi x',\S_S\circ \pi^{-1}) &= \begin{cases}
             \frac{N^k_{\pi T}}{|\pi S|^k} \S_S \circ \pi^{-1}(\pi x) & \text{if } \pi T\subseteq  \pi S\\
            0 & \text{if } \pi T\nsubseteq \pi S\\
        \end{cases}\\
    &= \begin{cases}
            \frac{N^k_{T}}{|S|^k} \S_S (x) & \text{if } T\subseteq  S\\
            0 & \text{if } T\nsubseteq S
        \end{cases}\\
    &= \text{RS}^k_{T}(x,x',\S_S)
\end{align*}
So RS is symmetric on synergy functions. Now use the synergy decomposition of $F\in \F$ to show RS is generally symmetric.

\subsubsection{Augmented Recursive Shapley and Properties}

The synergy function $\phi$ is taken implicitly with respect to a baseline appropriate to $F$. To make the baseline choice explicit, we write $\phi(F) =\phi(x',F)$. Augmented Recursive Shapley is then defined as:

\begin{equation}
    \text{RS}_T^{k*}(x,x',F) = \phi_T(x',F)(x) + \text{RS}_T^k(x,x', F- \sum_{S\in\pow_k}\phi_S(x',F))
\end{equation}
With the above augmentation, $\IH^{k*}$ explicitly distributes synergies $\phi_T(F)$ to group $T$ whenever $|T|\leq k$, and distributes higher synergies as $\IH^k$.

The above is a \textul{linear} function of $F$. Plugging in $\S_S$ to the above gains the following \textul{distribution policy}:
\begin{equation}\label{augmented recursive shap normal distribution}
    \text{RS}^{k*}_T(\S_S) = \begin{cases}
            \S_S(x) & \text{if } T=S\\
            \frac{N^k_T}{|S|^k}\S_S(x) & \text{if } T\subsetneq S, |S|>k\\
            0 & \text{else}
        \end{cases}
\end{equation}
Because each $F$ ha a unique synergy decomposition, we have

\begin{corollary}
    Augmented Recursive Shapley of order $k$ is the unique $\kth$-order interaction index that satisfies linearity and acts on synergy functions as in Eq.~\eqref{augmented recursive shap normal distribution}. 
\end{corollary}

To show that Augmented Recursive Shapley satisfies \textul{null feature}, let $F$ not vary in some feature $x_i$ and let $i\in T$. Then

\begin{align*}
    \text{RS}_T^{k*}(x, F) &= \sum_{S\in\pow_n}\text{RS}^{k*}_T(x,\phi_S(F))\\
    &= \text{RS}^{k*}_T(x,\phi_T(F)) + \sum_{T\subsetneq S, |S|>k}\text{RS}^{k*}_T(x,\phi_S(F))\\
    &= \text{RS}^{k*}_T(x,0) + \sum_{T\subsetneq S, |S|>k}\frac{N^k_T}{|S|^k}\phi_S(F)(x)\\
    &= 0 + \sum_{T\subsetneq S, |S|>k}0\\
    &= 0
\end{align*}
Thus Augmented Recursive Shapley satisfies null feature.

To show Augmented Recursive Shapley satisfies \textul{baseline test for interactions ($k\leq n$)}, let $T\subsetneq S$, $|S| \leq k$, and $\S_S\in C_S$. Then $\text{RS}_T^{k*}(x, \S_S)=0$ by Eq.\eqref{augmented recursive shap normal distribution}.

To show Augmented Recursive Shapley satisfies \textul{completeness}, consider the synergy function $\S_S$. If $|S|\leq k$, Eq.~\eqref{augmented recursive shap normal distribution} shows completeness. If $|S|>k$, then follow the proof of completeness for Recursive Shapley.

To show Augmented Recursive Shapley satisfies \textul{symmetry}, consider a synergy function $\S_S\in C_S$ and permutation $\pi$. Note that for $\S_S\in C_S$, we have $\S_S\circ\pi^{-1}$ is a pure interaction function in $\pi S$ with baseline $\pi x'$. Then

\begin{align*}
    \text{RS}^{k*}_{\pi T}(\pi x,\pi x',\S_S\circ \pi^{-1}) &= \begin{cases}
            \frac{N^k_{\pi T}}{|\pi S|^k} \S_S \circ \pi^{-1}(\pi x) & \text{if } \pi T =  \pi S\\
            \frac{N^k_{\pi T}}{|\pi S|^k}\S_S \circ \pi^{-1}(x) & \text{if } \pi T\subsetneq \pi S, |\pi S|>k\\
            0 & \text{else}\\
        \end{cases}\\
    &= \begin{cases}
            \frac{N^k_{T}}{|S|^k}\S_S (x) & \text{if } T\subseteq  S\\
            \frac{N^k_{T}}{|S|^k}\S_{S}(x) & \text{if } T\subsetneq S, |S|>k\\
            0 & \text{else} 
        \end{cases}\\
    &= \text{RS}^{k*}_{T}(x,x',\S_S)
\end{align*}

\subsection{Proof of Theorem~\ref{theorem: continuity condition}}\label{appendix: proof of continuity theorem for interactions}

\begin{proof}

    Let $\Int^k$ be a $\kth$-order interaction method defined for all $(x,x',F)\in[a,b]\times[a,b]\times\C$. Fix $x'$ and $x$. Let $T_l$ be the $l^\text{th}$ order Taylor approximation of $F$ at $x'$. Then

    \begin{align*}
        \Int^k(x,x',F)&= \lim_{l \rightarrow \infty}\Int^k(x,x',T_l)\\
        &=\sum_{m\in\N^n, \|m\|_1\leq l} \frac{D^m(F)(x')}{m!}\lim_{l \rightarrow \infty}\Int^k(x,x',(y-x')^m)
    \end{align*}
The last line is determined by the action of $\Int^k$ on elements of the set $\{(x,x',F): F(y) = (y-x')^m, m\in\N^n\}$, concluding the proof.

    






\end{proof}

\subsection{Proof of Corollary \ref{IG corollary}}\label{appendix: continuity condition theorem}
\cite{sundararajan2017axiomatic} has shown that IG is linear and Eq.~\eqref{IG_monomial_distribution} shows the actions of IG on polynomials.

Let $F\in\C$ and let $T_l$ be the Taylor approximation of $F$ of order $l$ centered at $x'$. It is known that $\frac{\partial T_l}{\partial x_i}\rightarrow \frac{\partial F}{\partial x_i}$ uniformly on a compact domain, such as $[a,b]$. Thus,
\begin{equation}
    \begin{split}
        \lim_{l\rightarrow \infty} \I_i(x,T_l) &= \lim_{l\rightarrow \infty} (x_i-x_i')\int_0^1 \frac{\partial T_l}{\partial x_i}(x'+t(x-x'))dt\\
        &= (x_i-x_i')\int_0^1 \frac{\partial F}{\partial x_i}(x'+t(x-x'))dt\\
        &= \I_i(x,F)
    \end{split}
\end{equation}
Thus IG satisfies the continuity criteria. Apply Theorem~\ref{theorem: continuity condition} for result.

\subsection{Integrated Hessian and Augmented Integrated Hessian}\label{IH_appendix}
\subsubsection{Definition of Integrated Hessian}
Here we give a complete definition of IH and detail how IH distributes monomials. We also detail $\IH^*$ and show it satisfies Corollary~$\ref{IH* corollary}$. We then show both satisfy completeness, linearity, null feature, and symmetry, and augmented IH satisfies baseline test for interactions ($k\leq n$).

Let $\sigma^k_T$ be the set of sequences of length $k$ such that the sequence is made of the elements of $T\neq \emptyset$ and each element appears at least once. For example, $\sigma_{\{1,2\}}^3 = \{(1,1,2),(1,2,1),(1,2,2),(2,1,1),(2,1,2),$ $(2,2,1)\}$. For a given sequence $s$, define $\IG_s(x,F)$ to be a recursive implementation of IG according to the sequence $s$, i.e., $\IG_{(1,2,3)}(x,F) = \IG_3(x,IG_2(\cdot,\IG_1(\cdot,F)))$.

We can then define the $k$-order Integrated Hessian for $T\neq \emptyset$ by:

\begin{equation}
    \IH^k_T(x,F) = \sum_{s\in\sigma^k_T}\IG_s(x,F),
\end{equation}
and for $T=\emptyset$, we define $\IH^k_\emptyset(x,x',F) = F(x')$.

\subsubsection{IH Policy Distributing Monomials and Continuity Condition}

We now move to inspect this equation and establish some properties. First, IG is linear, establishing that IH is also \textul{linear} by its form.

Next, we establish its \textul{policy distributing monomials} centred at $x'$. Eq.~\eqref{IG_monomial_distribution} states that for a monomial $F(y) = (y-x')^m$, $m\neq 0$,
\begin{equation}
    \IG_i(x,x',(y-x')^m) =
        \begin{cases}
            \frac{m_i}{\|m\|_1}(y-x')^m & \text{if } i \in S_m\\
            0 & \text{if } i \notin S_m
        \end{cases}
\end{equation}
Then for a given sequence $s\in\sigma^k_T$ and synergy function $(y-x')^m$, $T\subseteq  S_m$,

\begin{equation}
    \begin{split}
        \IG_s(x,(y-x')^m) &= \IG_{s_k}(x,\IG_{s_{k-1}}( ... \IG_{s_1}(\cdot, (y-x')^m)....)\\
        &= \IG_{s_k}(x,\IG_{s_{k-1}}(... \IG_{s_2}(\cdot, \frac{m_{s_1}(y-x')^m}{\|m\|_1})....)\\
        &= \IG_{s_k}(x,\IG_{s_{k-1}}(... \IG_{s_3}(\cdot, \frac{m_{s_1}m_{s_2}(y-x')^m}{\|m\|_1^2})....)\\
        &= ...\\
        &= \IG_{s_k}(x,\frac{\Pi_{1\leq i \leq k-1}m_{s_i}(y-x')^m}{\|m\|_1^{k-1}})\\
        &= \frac{\Pi_{1\leq i \leq k}m_{s_i}}{\|m\|_1^k}(x-x')^m\\
    \end{split}
\end{equation}
However, if there exists any elements of $s$ that is not in $S_m$, then:
\begin{equation}
        \begin{split}
        \IG_s(x,x',(y-x')^m) &= 0,
    \end{split}
\end{equation}
due to some ${s_j}\notin S_m$ in the sequence.

Now, applying the definition of IH when $m\neq 0$, we get:

\begin{equation}\label{IH and its distribution of Monomials}
    \begin{split}
        \IH^k_T(x,(y-x')^m) &= \sum_{s\in\sigma^k_T} \IG_s(x,(y-x')^m)\\
        &=\begin{cases}
            \sum_{s\in\sigma^k_T} \frac{\Pi_{1\leq i \leq k}m_{s_i}}{\|m\|_1^k}(x-x')^m & \text{if } T\subseteq  S_m\\
            \sum_{s\in\sigma^k_T} 0 & \text{if } T\nsubseteq S_m\\
        \end{cases}\\
        &= \begin{cases}
            \frac{M^k_T(m)}{\|m\|_1^k} (x-x')^m & \text{if } T\subseteq  S_m\\
            0 & \text{if } T\nsubseteq S_m,\\
        \end{cases}
    \end{split}
\end{equation}
where we define $M^k_T(m) = \sum_{|l|=k \text{ s.t. } S_l = T} {k \choose l}m^l$, with ${k \choose l} = \frac{k!}{\Pi_{i\in S_l} l_i!}$ the multinomial coefficient. In the case $T=S_m=\emptyset$, we set $\frac{M_T^k(m)}{\|m\|^k_1}=1$.

Now let us turn to the question of the \textul{continuity of Taylor approximation for analytic functions}. Let $T_l$ be the Taylor approximation of some $F\in\C$. Using Corollary~\ref{IG corollary}, we have $\lim_{l\rightarrow \infty} \IG_i(x,T_l) = \IG_i(x, F)$. This implies:

\begin{equation}
    \begin{split}
        \IG_i(x,F)  &= \lim_{l\rightarrow \infty} \IG_i(x,T_l)\\
        &= \sum_{m\in\N^n} \frac{D^m(F)(x')}{m!}\IG_i(x,(y-x')^m)\\
        &=\sum_{m\in\N^n}\frac{D^m(F)(x')}{m!}\frac{m_i}{\|m\|_1}(x-x')^m
    \end{split}
\end{equation}
That is, the above sum is convergent for all $x\in[a,b]$, implying that $\IG_i(\cdot,F)\in\C$. Also note:

\begin{equation}
    \begin{split}
        \IG_i(x,T_l) =\sum_{m\in\N^n, |m| \leq l}\frac{D^m(F)(x')}{m!}\frac{m_i}{\|m\|_1}(x-x')^m
    \end{split}
\end{equation}
This shows that $\IG(x,T_l)$ is a Taylor approximation of $\IG_i(x,F)$. Thus, for $F\in\C$ and a sequence $s$, we can pull the limit out consecutively since we are simply dealing with a series of Taylor approximations.
\begin{equation}
    \begin{split}
        \IG_s(x,F) &= \IG_{s_k}(x,\IG_{s_{k-1}}( ... \IG_{s_1}(\cdot, F)...))\\
        &= \IG_{s_k}(x,\IG_{s_{k-1}}( ... \lim_{l\rightarrow \infty}\IG_{s_1}(\cdot, T_l)...))\\
        &= \IG_{s_k}(x,\IG_{s_{k-1}}( ... \lim_{l\rightarrow \infty}\IG_{s_2}(\cdot,\IG_{s_1}(\cdot, T_l))...))\\
        &= \lim_{l\rightarrow \infty}\IG_{s_k}(x,\IG_{s_{k-1}}( ... \IG_{s_1}(\cdot, T_l)...))\\
        &= \lim_{l\rightarrow \infty} \IG_s(x, T_l),
    \end{split}
\end{equation}
which establishes that $\IH^k$ satisfies the continuity property. This implies the following corollary:

\begin{corollary}
    Integrated Hessian of order $k$ is the unique $\kth$-order method to satisfy linearity, the continuity condition, and distributes monomials as in Eq.~\eqref{IH and its distribution of Monomials}.
\end{corollary}

\subsubsection{Establishing Further Properties of IH}\label{section:establishing properties of IH}

To show IH is \textul{complete}, observe for a monomial $F(y) = (y-x')^m$, $m\neq 0$,

\begin{align*}
    \sum_{S\in\pow_k, |S|>0} \IH^k_S(x,x',F) &= \sum_{S\subseteq S_m, |S|>0} \frac{M^k_T(m)}{\|m\|_1^k} (x-x')^m\\
    &= \sum_{S\subseteq S_m, |S|>0} \frac{\sum_{|l|=k \text{ s.t. } S_l = S} {k \choose l}m^l}{\|m\|_1^k} (x-x')^m\\
    &= \frac{\|m\|_1^k}{\|m\|_1^k}(x-x')^m\\
    &= (x-x')^m
\end{align*}
When $m=0$, we get $\IH^k_S(x,x',F)=0$ except when $S=\emptyset$, in which case we get $\IH^k_S(x,x',F)=1$.

Applying the Taylor decomposition of $F$ and continuity property to a general $F\in\C$, we get:

\begin{align*}
    \sum_{S\in\pow_k, |S|>0} \IH^k_S(x,x',F) &= \sum_{S\in\pow_k, |S|>0} \lim_{l\rightarrow \infty} \IH^k_S(x,x',T_l)\\
    &= \lim_{l\rightarrow \infty}\sum_{S\in\pow_k, |S|>0}\sum_{m\in\N^n, 0<\|m\|_1\leq l} \frac{D^m(F)(x')}{m!} \IH^k_S(x,x',(y-x')^m)\\
    &= \lim_{l\rightarrow \infty}\sum_{m\in\N^n, 0<\|m\|_1\leq l} \frac{D^m(F)(x')}{m!} \sum_{S\in\pow_k, |S|>0}\IH^k_S(x,x',(y-x')^m)\\
    &= \lim_{l\rightarrow \infty}\sum_{m\in\N^n, 0<\|m\|_1\leq l} \frac{D^m(F)(x')}{m!} (x-x')^m\\
    &= \lim_{l\rightarrow \infty}\sum_{m\in\N^n, \|m\|_1\leq l} \frac{D^m(F)(x')}{m!} (x-x')^m - F(x')\\
    &= F(x) - F(x')
\end{align*}

To show IH satisfies \textul{null feature}, we proceed as in the proof for Recursive Shapley and suppose that $F$ does not vary in $x_i$. Then for any $S\in\pow_k$ such that $i\in S$, $\phi_S(F)=0$ since the synergy function is an interaction satisfying null feature. Then if $i\in T$,

\begin{equation}
    \begin{split}
        \IH_T^k(x,F) &= \sum_{S\in\pow_k} \IH_T^k(x,\phi_S(F))\\
         &= \sum_{S\in\pow_k \text{ s.t. } i\in S} \IH_T^k(x,\phi_S(F)) + \sum_{S\in\pow_k \text{ s.t. } i \notin S} \IH_T^k(x,\phi_S(F))\\
         &= \sum_{S\in\pow_k \text{ s.t. } i\in S} \IH_T^k(x,0) + \sum_{S\in\pow_k \text{ s.t. } i \notin S} 0\\
         &= 0
    \end{split}
\end{equation}

To show \textul{symmetry}, let $\pi$ be a permutation. Note that since $(\pi y)_{\pi(i)} = y_i$, we also have $(\pi^{-1} y)_i = (\pi^{-1} y)_{\pi^{-1} (\pi(i))} = y_{\pi(i)}$. Then, if $F(y) = (y-x')^m$, we get

\begin{align*}
    F\cdot \pi^{-1}(y) &= (y_{\pi(1)}-x_1')^{m_1}\cdot\cdot\cdot(y_{\pi(n)}-x_n')^{m_n}\\
    &= (y_1-x_{\pi^{-1}(1)}')^{m_{\pi^{-1}(1)}}\cdot\cdot\cdot(y_n-x_{\pi^{-1}(n)}')^{m_{\pi^{-1}(n)}}\\
    &= (y-\pi x')^{\pi m}
\end{align*}
Also note that,
\begin{align*}
    S_{\pi m} &= \{i: (\pi m)_i>0\}\\
    &= \{i: m_{\pi^{-1}(i)}>0\}\\
    &= \{\pi(i): m_{\pi^{-1}(\pi(i))}>0\}\\
    &= \{\pi(i): m_i>0\}\\
    &= \{\pi(i): i\in S_m\}\\
    &= \pi S_m
\end{align*}
Then,

\begin{align*}
    \IH^k_{\pi T}(\pi x,\pi x',F\circ \pi^{-1}) &= \begin{cases}
            \frac{M^k_{\pi T}(\pi m)}{\|\pi m\|_1^k} (\pi x-\pi x')^{\pi m} & \text{if } \pi T\subseteq  S_{\pi m}\\
            0 & \text{if } \pi T\nsubseteq S_{\pi m}\\
        \end{cases}\\
    &= \begin{cases}
            \frac{M^k_{T}(m)}{\|m\|_1^k} (x-x')^{m} & \text{if } T\subseteq  S\\
            0 & \text{if } T\nsubseteq S
        \end{cases}\\
    &= \IH^k_{T}(x,x',F)
\end{align*}

Now, if we take $\pi\in \C$ and denote $\pi^{-1}_j$ to be the $j^\text{th}$ output of $\pi^{-1}$, then $\frac{\partial \pi^{-1}_j}{\partial x_i} = \mathbbm{1}_{j = \pi^{-1}(i)}$. Then we have
\begin{align*}
    \frac{\partial (F\circ \pi^{-1})}{\partial x_i}(y) &= \sum_{j=1}^n \frac{\partial F}{\partial x_j}(\pi^{-1}(y)) \frac{\partial \pi_j^{-1}}{\partial x_i} (y)\\
    &= \frac{\partial F}{\partial x_{\pi^{-1}(i)}}(\pi^{-1}(y)),\\
\end{align*}
which yields
\begin{align*}
    D^{\pi m}(F\circ \pi^{-1})(\pi x')
    &= \frac{\partial^{\|\pi m\|_1} (F\circ \pi^{-1}) }{\partial x_1^{(\pi m)_1} \cdot\cdot\cdot\partial x_n^{(\pi m)_n} } (\pi x')\\
    &=\frac{\partial^{\|\pi m\|_1} F}{\partial x_{\pi^{-1}(1)}^{m_{\pi^{-1}(1)}}\cdot \cdot \cdot \partial x_{\pi^{-1}(n)}^{m_{\pi^{-1}(n)}}}(\pi^{-1}\pi x')\\
    &=\frac{\partial^{\|m\|_1} F}{\partial x_{1}^{m_{1}}\cdot \cdot \cdot \partial x_{n}^{m_{n}}}(x')\\
    &=D^{m}F(x')
\end{align*}
From the above we have for general $F$,
\begin{align*}
        \IH^k_{\pi S}(\pi x,\pi x',F\circ \pi^{-1}) &= \lim_{l\rightarrow \infty} \IH^k_{\pi S}(\pi x,\pi x', \sum_{m\in\N^n, 0<\|m\|_1\leq l} \frac{D^m(F\circ \pi^{-1})(\pi x')}{m!} (y-\pi x')^m)\\
        &= \lim_{l\rightarrow \infty}\sum_{m\in\N^n, 0<\|m\|_1\leq l} \frac{D^m(F\circ \pi^{-1})(\pi x')}{m!} \IH^k_{\pi S}(\pi x,\pi x',(y-\pi x')^m)\\
        &= \lim_{l\rightarrow \infty}\sum_{m\in\N^n, 0<\|m\|_1\leq l} \frac{D^{\pi m}(F\circ \pi^{-1})(\pi x')}{(\pi m)!} \IH^k_{\pi S}(\pi x,\pi x',(y-\pi x')^{\pi m})\\
        &= \lim_{l\rightarrow \infty} \sum_{m\in\N^n, 0<\|m\|_1\leq l} \frac{D^{ m}(F)(x')}{m!} \IH^k_{S}(x,x',(y-x')^{m})\\
        &= \lim_{l\rightarrow \infty}\IH_S(x,x',T_l)\\
        &= \IH_S(x,x',F)
\end{align*}

\subsubsection{Augmented Integrated Hessian and its Properties}

The synergy function $\phi$ is taken implicitly with respect to a baseline appropriate to $F$. To make the baseline choice explicit, we write $\phi(F) =\phi(x',F)$. Augmented Integrated Hessian is then defined as:

\begin{equation}\label{IH* equation}
    \IH_T^{k*}(x,x', F) = \phi_T(x',F)(x) + \IH_T^k(x,x', F- \sum_{S\in\pow_k}\phi_S(x',F))
\end{equation}
As in Augmented Recursive Shapley, Augmented Integrated Hessian explicitly distributes $\phi_T(F)$ to group $T$ when $|T|\leq k$, and distributes $\phi_T(F)$ as IH when $|T|>k$.

To establish the \textul{monomial distribution policy} we inspect the action of $\IH^{k*}_T$ in different cases. Plugging in $(y-x')^m$ to the above, if $|S_m|\leq k$, the right term is zero and Eq.~\eqref{IHk* Distribution Rule} holds, while if $|S_m|>k$, the left term is zero and the right term is $\IH_T^k(x,(y-x')^m)$. It is also easy to see that the above is \textul{linear}.

Regarding the \textul{continuity condition}, observe that:

\begin{align*}
    \phi_S(F) &= \sum_{m\in\N^n, S_m = S} \frac{D^{ m}(F)(x')}{m!} (x-x')^{m}\\
    &= \lim_{l\rightarrow \infty} \sum_{m\in\N^n, \|m\|_1\leq l, S_m = S} \frac{D^{ m}(F)(x')}{m!} (x-x')^{m}\\
    &= \lim_{l\rightarrow \infty} \phi_S (T_l),
\end{align*}
which gains,
\begin{align*}
    \lim_{l\rightarrow \infty} \IH_S^{k*}(x, T_l) &= \lim_{l\rightarrow \infty} \phi_S(T_l)(x) + \IH_S^k(x, T_l- \sum_{R\in\pow_k}\phi_R(T_l))\\
    &= \phi_S(F)(x) + \IH_S^k(x, \lim_{l\rightarrow \infty} T_l- \sum_{S\in\pow_k}\phi_R(T_l))\\
    &= \IH_S^k(x, F - \sum_{R\in\pow_k}\phi_R(F))\\
    &= \IH_S^{k*}(x, F),
\end{align*}
which establishes Corollary~\ref{IH* corollary}.

To show \textul{completeness}, consider the decomposition $F = \sum_{S\in\pow_n} \phi_S(F)$. Now $\IH^{k*}$ satisfies completeness for the subset of functions $\S_S\in C_S$, $|S|\leq k$ from the completeness of $\phi$ and Eq.~\eqref{IH* equation}. Also, $\IH^{k*}$ satisfies completeness for the subset of functions $\S_S\in C_S$, $|S|> k$ because $\IH^k$ satisfies completeness. From this we have:

\begin{align*}
    \sum_{T\in\pow_k, |T|\neq 0} \IH_T^{k*}(x,x',F) &= \sum_{T\in\pow_k, |T|\neq 0} \IH_T^{k*}(x,x', \sum_{S\in\pow_n} \phi_S(F))\\
    &= \sum_{S\in\pow_n} \sum_{T\in\pow_k, |T|\neq 0} \IH_T^{k*}(x,x', \phi_S(F))\\
    &= \sum_{S\in\pow_n, |S|\neq 0} [\phi_S(F)(x) - \phi_S(F)(x')]\\
    &= \sum_{S\in\pow_n, |S|\neq 0} [\phi_S(F)(x)] + F(x') - F(x')\\
    &= \sum_{S\in\pow_n} [\phi_S(F)(x)] - F(x')\\
    &= F(x) - F(x')
\end{align*}

\textul{Baseline test for interactions} applies immediately from the definition of Augmented Integrated Hessian in Eq.~\eqref{IH* equation}. Concerning \textul{null feature}, suppose $F$ does not vary in some $x_i$ and $i\in T$. First, we have $\phi_T(F)=0$. Also, $F-\sum_{R\in \pow_k}\phi_R(F)$ does not vary in $x_i$ either, so, since $\IH^k$ satisfies null feature. Thus we have $\IH^{k*}(x,F)=0$ by Eq.~\eqref{IH* equation}.

Lastly, concerning \textul{symmetry}, let $\pi$ be a permutation. Note that $\phi$ is symmetric, as it is the $k=n$ case for Shapley-Taylor, which is symmetric. Then,

\begin{align*}
    \IH^{k*}_{\pi T}(\pi x,\pi x',F\circ \pi^{-1}) &= \phi_{\pi T}(\pi x',F\circ \pi^{-1})(\pi x) + \IH_{\pi T}^k(\pi x,\pi x', F\circ \pi^{-1}- \sum_{R\in\pow_k}\phi_{\pi R}(\pi x',F\circ \pi^{-1}))\\
    &= \phi_{T}(x',F)(x) + \IH_{T}^k(\pi x,\pi x', \phi_{\pi R}(\pi x', \sum_{R\subset N, |R|>k} F\circ \pi^{-1}))\\
    &= \phi_{T}(x',F)(x) + \sum_{R\subset N, |R|>k} \IH_{T}^k(\pi x,\pi x', \phi_{\pi R}(\pi x', F\circ \pi^{-1}))\\
    &= \phi_{T}(x',F)(x) + \sum_{R\subset N, |R|>k} \IH_{T}^k(x,x', \phi_{R}(x',F))\\
    &= \phi_{T}(x',F)(x) + \IH_{T}^k(x,x', \sum_{R\subset N, |R|>k} \phi_{R}(x',F))\\
    &= \IH^{k*}_T(x,x',F)
\end{align*}

\subsection{Sum of Powers}\label{appendix:sum_of_powers}

\subsubsection{Defining Sum of Powers}
To define Sum of Powers, we first turn to defining a slight alteration of the Shapley-Taylor method. Suppose we performed Shapley-Taylor on a function $F$, but we treated $F$ as a function of every variable except for $x_i$, which we held at the input value. Specifically, for a given index $i$ and coalition $S$ with $i\in S$, we perform the $(|S|-1)$-order Shapley-Taylor method for the coalition $S\setminus\{i\}$. We perform this on an alteration of $F$, so that $F$ is a function of $n-1$ variables because the $x_i$ value is fixed. We denote this function $\ST^{-i}_S$, which has formula:

\begin{equation}
    \ST^{-i}_S(x,x',F) = \frac{|S|-1}{n-1}\sum_{T\subseteq  N\setminus S} \frac{\delta_{S\setminus \{i\}|T\cup\{i\}}F(x)}{{n-2 \choose |T|}}
\end{equation}

With this, we define Sum of Powers for $k\geq 2$ as:

\begin{equation}
    \SP^k_S(x,x',F) =
    \begin{cases}
        \sum_{i\in S}\left[\ST^{-i}_S(x,x',\IG_i(\cdot,x',F))\right ] & \text{if } |S|=k\\
        \phi_S(F) & \text{if } |S| < k\\
    \end{cases}
\end{equation}
We define the Sum of Powers for $k=1$ as the IG, with the addition that $\SP^1_\emptyset(x,x',F) = F(x')$.

Similar to the alteration of the Shapley-Taylor, we can alter the Shapley method, giving us:

\begin{equation}
    \Shap_j^{-i}(x,x',F) = \sum_{S\subset N\setminus \{i,j\}} \frac{|S|!(n-|S|-2)!}{(n-1)!} \left (F(x_{S\cup\{i,j\}}) - F(x_{S\cup\{i\}})\right)
\end{equation}
For the Sum of Powers $k=2$ case, the altered Shapley-Taylor is a $1$-order Shapley-Taylor method, and conforms to the Shapley method:
\begin{equation}
    \SP^2_{i,j}(x,x',F) =
    \begin{cases}
        \Shap_j^{-i}(x,x',\IG_i(\cdot,x',F)) + \Shap_i^{-j}(x,x',\IG_j(\cdot,x',F)) & \text{if } |S|=2\\
        \phi_S(F) & \text{if } |S| \leq 1\\
    \end{cases}
\end{equation}

\subsubsection{Proof of Corollary~\ref{Sum Power corollary}}
For the $k=1$ case, Sum of Powers is the IG, which satisfies linearity, distributes as in Eq.~\ref{SP distribution scheme}, and satisfies the continuity condition.

We now assume $k\geq 2$ for the rest of the section. First, $\SP^k_S$ satisfies \textul{linearity} because $\IG$ is linear in $F$ and $\ST_S^{-i}$ is linear in $F$.

We now proceed by cases to establish \textul{how $\SP^k$ distributes monomials}. We consider first the action of $\ST_S^{-i}$ on $F(y)=(y-x')^m$. $\ST_S^{-i}$ acts as the $(|S|-1)$-order Shapley-Taylor on an augmented function $F^{-i}(y_1,...,y_{i-1},y_{i+1},...,y_i) := (x_i-x_i')^{m_i}\Pi_{j\neq i}(y_j-x_j')^{m_j}$. Now, $\Pi_{j\neq i}(y_j-x_j')^{m_j}$ is a synergy function of $S_m\setminus\{i\}$. Thus we can use the distribution rule of Shapley-Taylor, gaining

\begin{equation}
\begin{split}
    \ST_S^{-i}(x,x',F) &= \ST_{S\setminus\{i\}}^{|S|-1}(x_{-i}, x'_{-i},F^{-i})\\
    &= \begin{cases}
     (x_i-x_i')^m & \text{if } S=S_m\\
     \frac{(x_i-x_i')^m}{{|S|-1\choose k-1}} & \text{if } S\subsetneq S_m, |S|=k\\
     0 & \text{else}
 \end{cases},
\end{split}
\end{equation}
where $x_{-i}$ denotes the vector $x$ with the $i^\text{th}$ component removed.

With this established, we now show the action of the Sum of Powers method for an exhaustive set of cases:

\begin{enumerate}
    \item ($|S|<k$, $S=S_m$): $\SP_S^k(x,(y-x')^m) = \phi_S((y-x')^m) = (y-x')^m$.
    \item ($|S|<k$, $S\neq S_m$): $\SP_S^k(x,(y-x')^m) = \phi_S((y-x')^m) = 0$.
    \item($|S|=k$, $S\subseteq  S_m$):
\begin{equation}\nonumber
    \begin{split}
        \SP_S^k(x,x', (y-x')^m) &= \sum_{i\in S}\left[\ST^{-i}_S(x,x',\IG_i(\cdot,x',(y-x')^m)\right ]\\
        &= \sum_{i\in S}\left[\ST^{-i}_S(x,x',\frac{m_i}{\|m\|_0}(y-x')^m)\right ]\\
        &= \sum_{i\in S}\frac{1}{{|S_m|-1 \choose |S|-1}} \frac{m_i}{\|m\|_0}(x-x')^m\\
        &= \frac{1}{{|S_m|-1 \choose |S|-1}} \frac{\sum_{i\in S}m_i}{\|m\|_0}(x-x')^m
    \end{split}
\end{equation}

    \item ($|S|=k$, $S\nsubseteq S_m$): Let $i\in S$. If $i\in S\setminus S_m$, then $\ST^{-i}_S(x,x',\IG_i(\cdot,x',(y-x')^m)) = \ST^{-i}_S(x,x',0)) = 0$.
    
    If, on the other hand, $i\in S_m$, then $\ST^{-i}_S(x,x',\IG_i(\cdot,x',(y-x')^m)) = \ST^{-i}_S(x,x',\frac{m_i}{\|m\|_0}(y-x')^m)$. Now, the altered Shapley-Taylor takes the value of zero for synergy functions of sets that are not super-sets of the attributed group, $S\setminus\{i\}$. Also, $(y-x')^m$ is a synergy function of $S_m$, and $S_m$ is not a super-set of $S\setminus\{i\}$. Thus $\ST^{-i}_S(x,x',\frac{m_i}{\|m\|_0}(y-x')^m)=0$.

    This established that each term in the sum $\sum_{i\in S}\left[\ST^{-i}_S(x,x',\IG_i(\cdot,x',(y-x')^m))\right ]$ is zero, gaining $\SP^k_S(x,x',(y-x')^m=0$.
\end{enumerate}

Thus Sum of Powers has a distribution scheme that agrees with Eq.~\eqref{SP distribution scheme}. To restate:

\begin{align}
    \SP^k_T(x,(y-x)^m) =
    \begin{cases}
        (x-x')^m & \text{if } T=S_m\\
        \frac{1}{{|S_m|-1 \choose k-1}} \frac{\sum_{i\in T}m_i}{\|m\|_1}(x-x')^m & \text{if } T\subsetneq S_m, |T|=k\\
        0 & \text{else}
    \end{cases}
\end{align}

Finally, IG satisfies the continuity condition by Corollary~\ref{IG corollary}, and it is easy to see that that $\ST^{-1}_S$ satisfies the continuity condition. Thus Sum of Powers obeys the \textul{continuity condition}. 

\subsubsection{Establishing Further Properties for Sum of Powers}
To establish \textul{null feature}, let $F$ not vary in $x_i$ and let $i\in S$. Sum of Powers satisfies the continuity condition, so

\begin{align*}
    \SP^k_S(x,x',F) &= \lim_{l\rightarrow \infty} \sum_{m\in\N^n, |m| \leq l} \frac{D^m F(x')}{m!} \SP^k_S(x,x',(y-x')^m)\\
    &=\lim_{l\rightarrow \infty} \sum_{m\in\N^n, |m| \leq l, m_i=0} \frac{D^m F(x')}{m!} \SP^k_S(x,x',(y-x')^m)\\
    &=0,\\
\end{align*}
where the second line is because $D^m F(x')=0$ if $m_i>0$ because $F$ does not vary in $x_i$, and the third line is because $\SP^k_S(x,x',(y-x')^m)=0$ if $m_i = 0$.

To establish \textul{baseline test for interaction ($k\leq n$)}, let $\S_S\in\C$ be a synergy function of $S$ and let $T\subsetneq S$, $|T|<k$. Then $\SP^k_T(x,\S_S) = \phi_T(\S_S)(x) = 0$.

To establish \textul{completeness}, consider $F(y)=(y-x')^m$, with $|S_m|>k$. Then,

\begin{align*}
    \sum_{S\in \pow_k, |S|>0} \SP^k_S(x,x',F) &= \sum_{S\subsetneq S_m, |S|=k} \SP^k_S(x,x',(y-x')^m)\\
    &=\sum_{S\subsetneq S_m, |S|=k} \frac{1}{{|S_m|-1 \choose k-1}} \frac{\sum_{i\in S}m_i}{\|m\|_1}(x-x')^m\\
    &= \frac{(x-x')^m}{{|S_m|-1 \choose k-1}\|m\|_1}\sum_{S\subsetneq S_m, |S|=k}\sum_{i\in S} m_i\\
    &= \frac{(x-x')^m}{{|S_m|-1 \choose k-1}\|m\|_1}{|S_m|-1 \choose k-1}\|m\|_1\\
    &=F(x) - F(x')
\end{align*}
Now treating a general $F\in \C$, the proof is identical to the proof for Integrated Hessian,

\begin{align*}
    \sum_{S\in\pow_k, |S|>0} \SP^k_T(x,x',F) &= \sum_{S\in\pow_k, |S|>0} \lim_{l\rightarrow \infty} \SP^k_T(x,x',T_l)\\
    &= \lim_{l\rightarrow \infty}\sum_{S\in\pow_k, |S|>0}\sum_{m\in\N^n, 0<\|m\|_1\leq l} \frac{D^m(F)(x')}{m!} \SP^k_T(x,x',(y-x')^m)\\
    &= \lim_{l\rightarrow \infty}\sum_{m\in\N^n, 0<\|m\|_1\leq l} \frac{D^m(F)(x')}{m!} \sum_{S\in\pow_k, |S|>0}\SP^k_T(x,x',(y-x')^m)\\
    &= \lim_{l\rightarrow \infty}\sum_{m\in\N^n, 0<\|m\|_1\leq l} \frac{D^m(F)(x')}{m!} (x-x')^m\\
    &= \lim_{l\rightarrow \infty}\sum_{m\in\N^n, \|m\|_1\leq l} \frac{D^m(F)(x')}{m!} (x-x')^m - F(x')\\
    &= F(x) - F(x')
\end{align*}

To show \textul{symmetry}, the proof parallels the proof for Integrated Hessian in section~\ref{section:establishing properties of IH}. Let $\pi$ be a permutation. If we let $F(y) = (y-x')^m$ and follow what was previously established in section~\ref{section:establishing properties of IH}, then

\begin{align*}
    \SP^k_{\pi T}(\pi x,\pi x',F\circ \pi^{-1}) &= \begin{cases}
            (\pi x-\pi x')^{\pi m} & \text{if } \pi T=S_{\pi m}\\
            \frac{1}{{|S_{\pi m}|-1 \choose k-1}} \frac{\sum_{i\in \pi T}(\pi m)_i}{\|\pi m\|_1}(\pi x-\pi x')^{\pi m} & \text{if } \pi T\subsetneq S_{\pi m}, |\pi T|=k\\
            0 & \text{else}
    \end{cases}\\
    &=  \begin{cases}
            (x-x')^m & \text{if } T=S_m\\
            \frac{1}{{|S_m|-1 \choose k-1}} \frac{\sum_{i\in T}m_i}{\|m\|_1}(x-x')^m & \text{if } T\subsetneq S_m, |T|=k\\
            0 & \text{else}
    \end{cases}\\
    &= \SP^k_{T}(x,x',F)
\end{align*}
From the above we have for general $F$,
\begin{align*}
        \SP^k_{\pi S}(\pi x,\pi x',F\circ \pi^{-1}) &= \lim_{l\rightarrow \infty} \SP^k_{\pi S}(\pi x,\pi x', \sum_{m\in\N^n, 0<\|m\|_1\leq l} \frac{D^m(F\circ \pi^{-1})(\pi x')}{m!} (y-\pi x')^m)\\
        &= \lim_{l\rightarrow \infty}\sum_{m\in\N^n, 0<\|m\|_1\leq l} \frac{D^m(F\circ \pi^{-1})(\pi x')}{m!} \SP^k_{\pi S}(\pi x,\pi x',(y-\pi x')^m)\\
        &= \lim_{l\rightarrow \infty}\sum_{m\in\N^n, 0<\|m\|_1\leq l} \frac{D^{\pi m}(F\circ \pi^{-1})(\pi x')}{(\pi m)!} \SP^k_{\pi S}(\pi x,\pi x',(y-\pi x')^{\pi m})\\
        &= \lim_{l\rightarrow \infty} \sum_{m\in\N^n, 0<\|m\|_1\leq l} \frac{D^{ m}(F)(x')}{m!} \SP^k_{S}(x,x',(y-x')^{m})\\
        &= \lim_{l\rightarrow \infty}\SP(x,x',T_l)\\
        &= \SP(x,x',F)
\end{align*}

\end{document}